\documentclass[final,5p,times,twocolumn]{elsarticle}

\usepackage{amssymb}
\usepackage{amsmath}
\usepackage{booktabs}
\usepackage{array}
\usepackage{multirow}
\usepackage{hyperref}
\usepackage{listings}
\usepackage{xcolor}
\usepackage[ruled,vlined]{algorithm2e}
\usepackage{tabularx}
\usepackage{listings}
\makeatletter\def\ps@pprintTitle{\let\@oddfoot\@empty\let\@evenfoot\@empty}\makeatother
\begin{document}

\begin{frontmatter}



\title{Joint Optimization of Storage and Loading for High-Performance 3D Point Cloud Data Processing}


\author{Ke Wang}
\author{Yanfei Cao}
\author{Xiangzhi Tao}
\author{Naijie Gu}
\author{Jun Yu}
\author{Zhengdong Wang}
\author{Shouyang Dong}
\author{Fan Yu}
\author{Cong Wang}
\author{Yang Luo}


\begin{abstract}
With the rapid development of computer vision and deep learning, significant advancements have been made in 3D vision, particularly in autonomous driving, robotic perception, and augmented reality. 3D point cloud data, as a crucial representation of 3D information, has gained widespread attention. However, the vast scale and complexity of point cloud data present significant challenges for loading and processing and traditional algorithms struggle to handle large-scale datasets.The diversity of storage formats for point cloud datasets (e.g., PLY, XYZ, BIN) adds complexity to data handling and results in inefficiencies in data preparation. Although binary formats like BIN and NPY have been used to speed up data access, they still do not fully address the time-consuming data loading and processing phase. To overcome these challenges, we propose the .PcRecord format, a unified data storage solution designed to reduce the storage occupation and accelerate the processing of point cloud data. We also introduce a high-performance data processing pipeline equipped with multiple modules. By leveraging a multi-stage parallel pipeline architecture, our system optimizes the use of computational resources, significantly improving processing speed and efficiency. This paper details the implementation of this system and demonstrates its effectiveness in addressing the challenges of handling large-scale point cloud datasets.On average, our system achieves performance improvements of 6.61x (ModelNet40), 2.69x (S3DIS), 2.23x (ShapeNet), 3.09x (Kitti), 8.07x (SUN RGB-D), and 5.67x (ScanNet) with GPU and 6.9x, 1.88x, 1.29x, 2.28x, 25.4x, and 19.3x with Ascend.
\end{abstract}

%


\end{frontmatter}



\section{Introduction}
\label{sec1}

With the rapid advancements in computer vision and deep learning, the field of 3D vision has achieved remarkable progress. As a fundamental representation of 3D data, point cloud data has been extensively applied in key domains such as autonomous driving, robotic perception, augmented reality (AR), and virtual reality (VR)\cite{xu2019autonomous}. Comprising a vast number of 3D coordinate points, point clouds accurately capture the geometric structure of objects or scenes, offering unique advantages in representing spatial configurations and object shapes in complex environments. Unlike traditional image data, point clouds not only provide precise surface and shape information but also encode depth and fine-grained variations\cite{guo2020deep}, making them indispensable for tasks such as 3D reconstruction\cite{xu2024cp,wang2021cgnet},3D scene completion, 3D object detection and tracking\cite{zhu2022vpfnet},3D shape classification\cite{hu2022decouple,hu2024novel,guo2023novel} and semantic segmentation\cite{zhang2024ppdistiller,lai2022tackling,zhou2023sampling,cheng2021ptanet,zhang2024pointgt}.

Despite their utility, the high-dimensional nature and intrinsic complexity of point cloud data pose significant challenges for storage and processing. Point cloud datasets, such as ScanNet\cite{dai2017scannet} and KITTI\cite{geiger2015kitti}, typically comprise millions of points and exhibit substantial spatial non-uniformity, creating pronounced disparities in the computational difficulty of processing sparse versus dense regions. Traditional algorithms often struggle to efficiently handle large-scale point clouds, especially in real-time applications requiring instantaneous feedback. For example, in autonomous driving\cite{xu2019autonomous,li2020lidar}, the ability to quickly read and analyze point cloud data is critical for ensuring safety. Delays in data processing can directly impact system response times, potentially compromising vehicle decision-making and control\cite{li2020lidar}. Thus, enhancing the efficiency of point cloud data loading, storage, and analysis is crucial for advancing 3D vision systems and ensuring their reliability.

The development of point cloud data has undergone multiple stages. To facilitate applications in various fields, point cloud datasets are stored in diverse formats, like PLY\cite{turk1994ply}, obj, XYZ, etc, which often necessitate format conversion before experiments. Such conversions increase implementation complexity and consume substantial time, particularly when dealing with large-scale data. While some new formats like BIN or NPY\cite{krijnen2017ifc} have been adopted to accelerate data access, these methods fail to address the bottlenecks associated with data loading, especially during data preparation, which often represents a major drain on system resources. 

Large-scale point cloud datasets also demand considerable storage capacity, posing challenges for both storage and transmission\cite{li2022plenoptic}. For instance, the KITTI\cite{geiger2015kitti} dataset occupies 42.9GB, the SUN RGB-D\cite{song2015sun} dataset requires 72.4GB, and the ScanNet V2\cite{dai2017scannet} dataset reaches an enormous 1.2TB. Such datasets not only require substantial local storage but also impose significant demands on I/O performance during loading and processing\cite{golla2015real,cao20193d}. These challenges are further exacerbated in cloud-based processing environments, where storage limitations can hinder the full utilization of point cloud datasets, leading to interruptions in training and inference tasks.

Apart from storage challenges, the efficiency of data loading and processing also plays a critical role in point cloud applications. Most point cloud-related 3D tasks rely on Python-based deep learning frameworks, such as PyTorch\cite{paszke2019pytorch} and TensorFlow\cite{abadi2015tensorflow}, for data loading and processing. While these frameworks offer parallel computing capabilities, such as multi-process data loading and batch processing, they encounter limitations when processing large-scale datasets. Despite leveraging GPUs and TPUs for computation\cite{jouppi2017datacenter}, data processing and loading stages, particularly for point clouds, remain bottlenecks. Inefficient utilization of CPU and memory resources during these stages adversely impacts overall system performance.

To address these challenges, we propose a high-performance solution consisting of four key modules. First, we introduce a high-efficiency data storage format, \textbf{PcRecord}, designed to standardize storage conventions across point cloud datasets. This format reduces storage requirements and significantly improves data access speed, simplifying large-scale data handling and processing. The structure of \textbf{PcRecord} is detailed in Section 3.Second, we present a cloud-based streaming dataset loading method using OBS\cite{ranjan2014streaming}, which enables scalable and on-demand processing. By storing datasets in the cloud and dynamically loading data during execution, we avoid the need to load entire datasets into memory. This streaming mechanism supports real-time, distributed processing across multiple devices\cite{hongchao2011distributed}, improving data handling efficiency.Third, we propose a high-performance data processing pipeline built on the MindSpore deep learning framework\cite{huawei2022huawei,chen2021deep} . This pipeline uses multi-stage parallelism to concurrently execute data processing tasks, significantly enhancing throughput.Finally, we incorporate an automatic tuning (autotune) strategy that dynamically optimizes key parameters\cite{bergstra2012random}, such as data loading and processing configurations, based on hardware and dataset characteristics. This approach not only adapts to diverse hardware environments but also ensures optimal performance across experimental and production scenarios.

The primary contributions of this paper are as follows: \begin{enumerate} 
\item \textbf{Proposed a novel system}: We introduce a novel dataset loading and processing system catering to large-scale point cloud dataset.
\item \textbf{Designed an efficient file format}: We design an efficient point cloud data storage format \textbf{.PcRecord}, which significantly reduces storage space of point cloud through optimized  compression algorithm and data management.
\item \textbf{Introduced a high-performance pipeline}: We present a data loading and processing pipeline along with \textbf{distributed processing} and autotune algorithm, which greatly boost the throughout rate of handling point cloud.
\item \textbf{Utilized OBS to minimize local storage}: We introduce OBS technology to prevent point cloud data from occupying large amounts of local storage space. 
\item \textbf{Outperformed mainstream deep learning frameworks}: The point cloud loading and processing performance of our system significantly outperforms mainstream deep learning frameworks including PyTorch, Tensorflow, Keras, MindSpore.

\end{enumerate}

This work addresses key bottlenecks in point cloud data processing and offers a promising path for future research. With continuous advancements in hardware and deep learning frameworks, efficient point cloud data processing will undoubtedly remain a critical focus in both research and industrial applications.

\begin{figure*}[!h]
    \centering
    \includegraphics[width=1\linewidth]{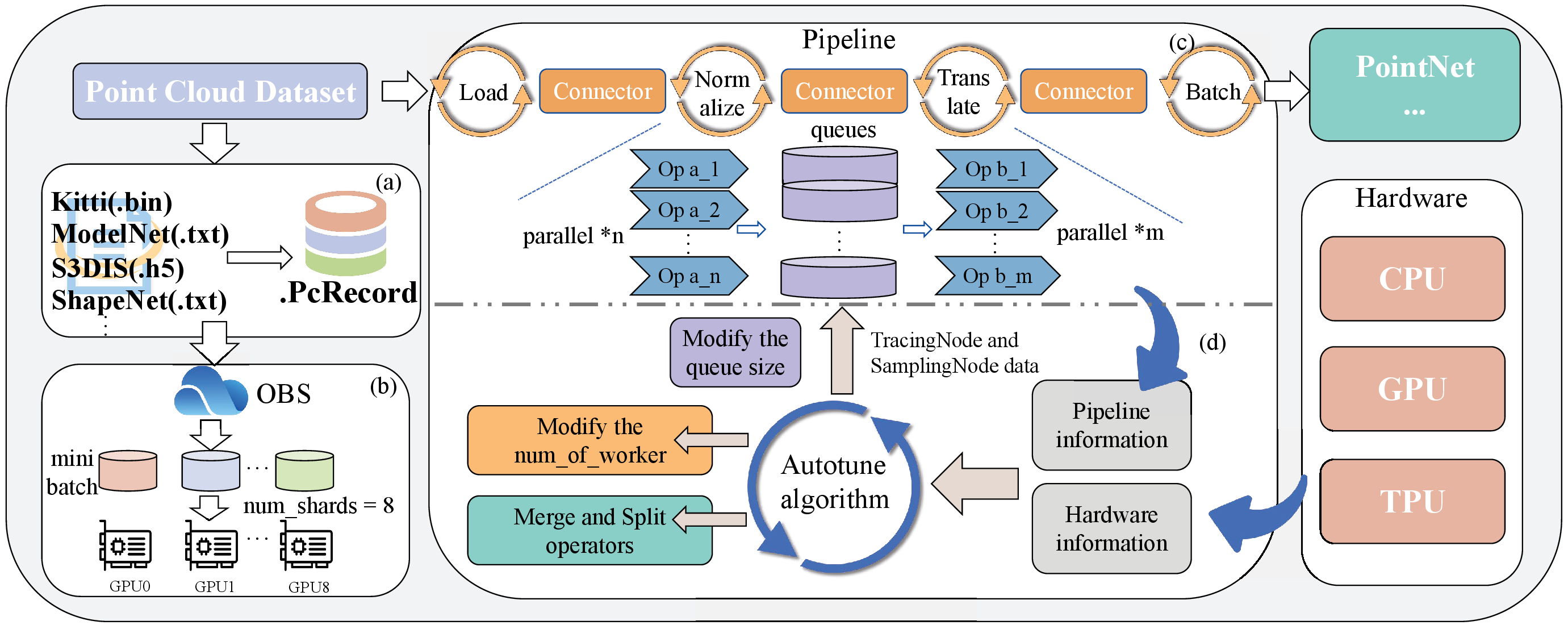}
    \caption{Overview of the proposed point cloud data processing pipeline with (a) Format Standardization
 Module, (b) Distributed Processing  Module, (c)Data Loading and Processing Pipeline, and (d) Autotune Module.}
    \label{fig:enter-label}
\end{figure*}

\section{Related work}
Point cloud data can be stored in various formats, with researchers and engineers developing tailored methods to meet diverse application requirements and scenarios. Traditional storage formats, such as OBJ, PLY, and PCD\cite{turk1994ply}, have distinct characteristics and application domains\cite{mchenry2008overview}.
\subsection{Most common storage formats}
\textbf{OBJ Format:} The OBJ format, developed by Wavefront as a general-purpose 3D model file format, is widely adopted in 3D visualization. It supports the representation of points, normals, texture coordinates, and other elements, typically in an ASCII text format. While this format is human-readable, it is less storage-efficient when handling large-scale point cloud data due to its verbosity.

\textbf{PLY Format:} The Polygon File Format (PLY), introduced by Turk et al. at Stanford University, was designed for storing and exchanging 3D model data. PLY supports both ASCII and binary storage modes, with the binary option significantly reducing storage requirements and enhancing read/write speeds. As a result, this format is extensively utilized in academia and research, particularly for representing point clouds and 3D scanned data.

\textbf{PCD Format:} The Point Cloud Data (PCD) format, developed by the Point Cloud Library (PCL), is specifically designed for storing and processing 3D point cloud data. It supports diverse data types, including floating-point numbers, RGB color attributes, and intensity values. This format is widely applied in fields such as computer vision and robotic perception, becoming a standard format in both industry and academia.

In addition to these specialized formats, simpler \textbf{text-based formats} like XYZ—which stores 3D coordinates in plain text—and CSV or TXT files are also commonly used. These formats are straightforward and suitable to feed neural networks during training. However, their lack of compression and optimization often results in large file sizes and substantial storage demands, especially for large-scale datasets.

\subsection{Most common methods of processing point cloud}

To address the challenges associated with the storage, loading and processing of point cloud data, numerous open-source libraries have emerged. 

\textbf{The Point Cloud Library} (PCL) is a prominent open-source project that provides a comprehensive suite of tools for processing 2D/3D images and point cloud data. Its functionalities include filtering, feature extraction, registration, segmentation, and surface reconstruction, making it a key resource for applications in robotics, autonomous driving, and computer vision.

\textbf{Open3D}, another widely used open-source library, focuses on the efficient development and processing of 3D data\cite{zhou2018open3d}. Built on C++ with a Python interface, it integrates point cloud processing, 3D geometric operations, and deep learning model interfaces, establishing itself as a versatile tool in 3D vision research and development.

\subsection{Deep learning-based point cloud tasks}

The rise of deep learning has further revolutionized the processing of raw point cloud data by bypassing traditional projection and discretization methods. A seminal contribution in this area is \textbf{PointNet}\cite{qi2017pointnet++}, which introduced a neural network architecture designed specifically for point clouds. 

PointNet utilizes symmetric functions, such as max pooling, to achieve invariance to point permutations, enabling the direct processing of raw point cloud data. It demonstrated significant success in 3D classification and segmentation tasks, achieving state-of-the-art performance on datasets like S3DIS and laying the groundwork for subsequent advances in deep learning-based point cloud processing. Like \textbf{PointNet}, \textbf{PointCNN}\cite{li2018pointcnn} introduced convolutional operations tailored to point cloud data, facilitating more efficient local feature extraction. These DL methods both employ binary storage formats such as .h5, which provide efficient access and are particularly suitable for neural network training.

As deep learning frameworks increasingly support the direct processing of raw point cloud data, binary formats have gained prominence for their ability to optimize loading and processing efficiency.

For example, the \textbf{NumPy} (.npy) format\cite{oliphant2006guide} is widely used in Python for storing large-scale array data. Its platform compatibility and storage efficiency make it a popular choice for research and open-source projects involving 3D datasets. Additionally, the \textbf{pickle} format\cite{python2017pickle} enables the serialization of Python objects into binary data, facilitating efficient data transmission and sharing across computational environments.

Leading deep learning frameworks, such as PyTorch and TensorFlow, have also advanced their data processing capabilities to accommodate large-scale datasets. PyTorch’s DataLoader optimizes data processing through multi-threading and batch loading strategies. However, I/O bottlenecks remain a challenge for extremely large datasets, despite the adoption of multi-threading and caching techniques. Similarly, TensorFlow’s tf.data API\cite{murray2021tf} provides a flexible and efficient pipeline for batching, preprocessing, and augmentation. While it excels in handling traditional image data, its performance in processing high-dimensional and irregular point cloud data is constrained by the limitations of existing data storage formats and read speeds.

\subsection{Processing of point cloud data in real-world applications}
In addition to the general challenges of processing large-scale point cloud datasets, certain application domains, such as autonomous driving (AV), introduce unique requirements that further complicate data storage and processing tasks.In the field of AV, point cloud data is often generated by LiDAR sensors to create a real-time 3D map of the vehicle's surroundings. The large volume of point cloud data produced by LiDAR sensors—combined with the need for high accuracy and low latency—requires highly optimized data storage and processing solutions. 

Traditional storage formats and deep learning frameworks often struggle to meet these requirements, as they may encounter bottlenecks when handling large, unstructured datasets in real-time environments. To meet the performance requirements of high throughput and real-time processing, many approaches have been proposed.

NVIDIA has launched the autonomous driving platform, NVIDIA DRIVE AGX, which is equipped with specially designed hardware DRIVE Thor and software DriveOS with TensorRT for real-time AI inference to accelerate deep learning computation. This platform enables autonomous vehicles to efficiently process sensor data, including point cloud data and image data, by supporting parallel computing and real-time inference.

Waymo from Google optimizes its system’s performance by employing heterogeneous computing platforms, including AI accelerators like TPUs, to speed up deep learning model training and inference. Distributed computing further optimizes performance by offloading heavy tasks to the cloud, while edge devices handle real-time processing on the vehicle. Additionally, Waymo uses techniques like model quantization and compression to minimize memory usage and computational load without compromising accuracy. Incremental data loading and optimized data pipelines ensure fast and efficient data retrieval and processing, preventing bottlenecks and improving overall system performance.

\section{Methods}

Fig. 1 provides an overview of the proposed system. After users input point cloud dataset into the system, the format standardization module (a) converts the data from any raw format into the unfied .PcRecord format. If the user chooses to activate streaming dataset loading approach (b), the module will send a download request to the OBS and initiate the data loading and processing pipeline once the download is complete. Also, if the user chooses to activate distributed processing approach (b), the module will ask the user to set the value of num\_shards (like 8).Then the module partitions the point cloud data and distributes it across multiple devices for concurrent processing. Each device independently launches a process to handle its assigned portion of the data within its own pipeline.

\begin{figure}[!h]
    \centering
    \includegraphics[width=1\linewidth]{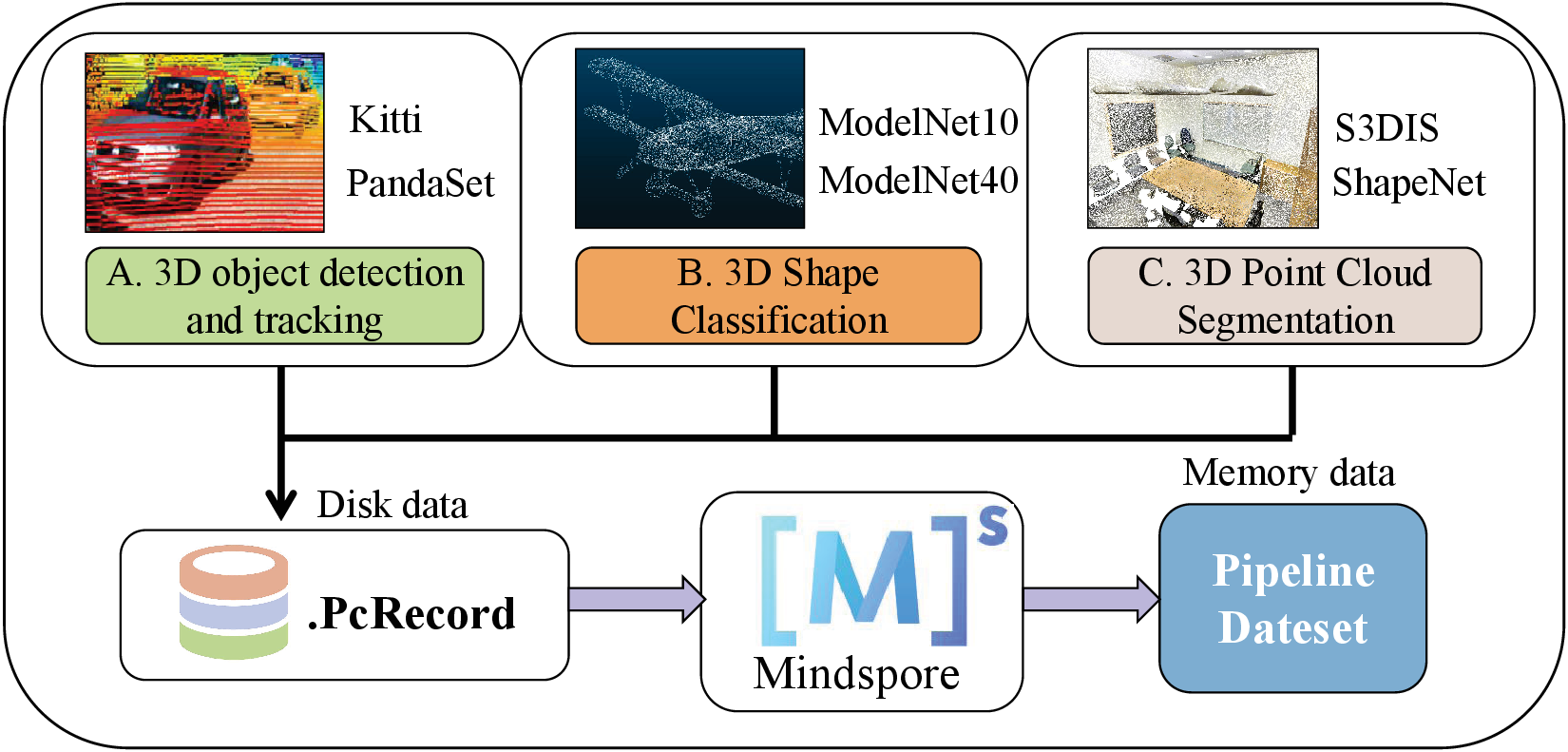}
    \caption{Workflow of the format standardization module.}
    \label{fig:enter-label}
\end{figure}

Next, the point cloud data undergoes processing through a multi-stage parallel pipeline (c). This pipeline efficiently handles parallel tasks while maintaining data order consistency using a polling algorithm. Upon initiation, the autotune strategy module (d) activates a monitoring thread that collects a wide range of information from both the hardware and the pipeline. The module treats the parallelism of the data processing pipeline as a hyperparameter optimization problem. By gathering real-time execution time and resource usage metrics, and utilizing a combination of random search and Bayesian optimization, it iteratively adjusts the pipeline to achieve optimal performance, striking a balance between pipeline efficiency and hardware resource utilization.

Our data processing engine, built on the open-source MindSpore framework, significantly enhances the performance of MindSpore, surpassing several leading deep learning frameworks in terms of point cloud data loading and processing efficiency.

\subsection{Format standardization module}
To address the issues of  various storge formats, large storage space usage, low reading efficiency, and reliance on external libraries, we have converted point cloud datasets into a new unified storage format, \textbf{.PcRecord} and use our system's unified API interface to load it with Mindspore, as shown in Fig 2.
This new efficient data format can significantly reduce disk I/O and network I/O overhead(Section 4.2 confirm it). The structure of the new data storage format is shown in Fig 3.
\begin{figure}[!h]
    \centering
    \includegraphics[width=0.75\linewidth]{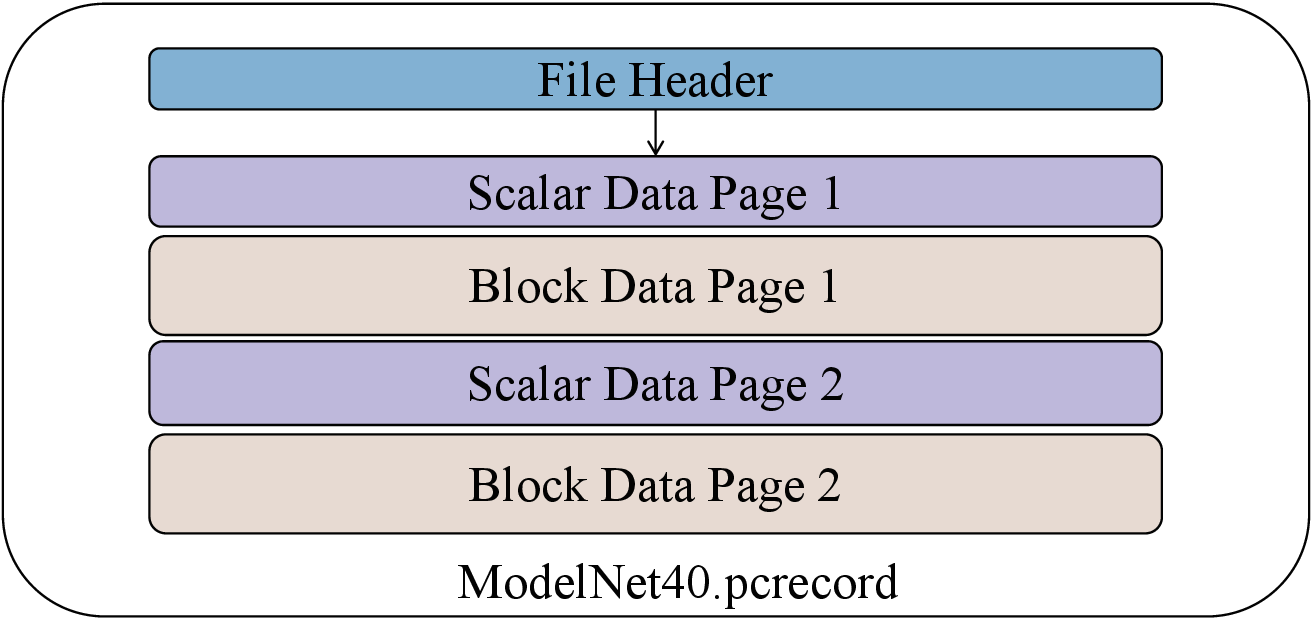}
    \caption{The structure of .PcRecord.}
    \label{fig:enter-label}
\end{figure}
The data file contains a "File Header", which records the size of the converted file in bytes, as well as the byte size of each Scalar Data Page and Block Data Page. The header also includes a Schema that describes the fields contained in the new file format and the type of each field in dictionary, which users should write by themselves before the conversion. 

For example, for the ModelNet40 dataset, its original format is divided into multiple folders named after object types (like airplane, table). Each file contains point cloud data of a specific object in binary format, which include basic point cloud information like x, y, z coordinates and the normal vector coordinates. So the Schema in file header for ModelNet40 will be like Fig 4. 
\begin{figure}[h]
    \centering
    \includegraphics[width=1\linewidth]{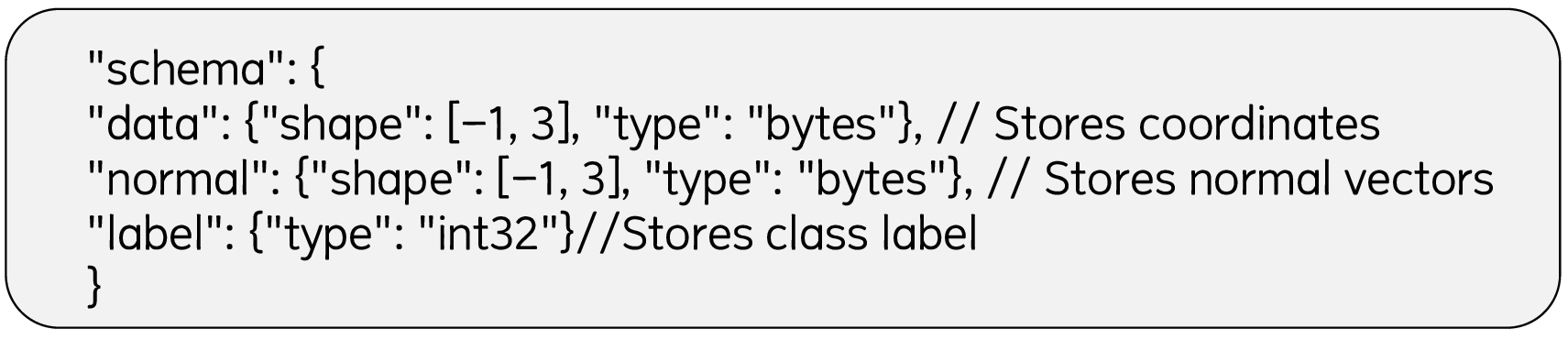}
    \caption{The schema of ModelNet40 dataset.}
    \label{fig:enter-label}
\end{figure}
It means ModelNet40 has 3 fileds: “data”, “normal” and “label”. “Data” and “normal” has 3 dimensions and are in bytes type while “Label” is int32 type. 

As for the second section "Scalar Data Page" in Fig 3, it primarily stores non-binary data types. For ModelNet40, it will include category labels such as "airplane" ,"table" using numerical identifiers.

The third section, “Block Data Page”, is mainly used to store binary data. So the coordinates and normal vector information of the ModelNet40 will be stored here. Though stored dividually, the data in Scalar Data Page and Block Data Page should be correspond to each other.

Besides the novel data storage and management method, .PcRecord also introduces a method specifically designed for compressing point cloud datasets's storage space. To be specific, .PcRecord adopts a dual-layer mechanism for data page compression. The outer layer applies the LZ4 compression algorithm at the page level, which works by using a sliding window to find repeating patterns, offering fast compression and decompression speeds. While the inner layer applies a column-specific optimization tailored for point cloud data. Given that point cloud datasets are of composed of coordinates and normal vectors, the inner layer employs differential encoding for coordinates, where only the differences between successive points are stored to reduce the overall storage size.  This hybrid approach optimizes both storage efficiency and access speed by balancing compression ratio with computational overhead. Benefited from the dual-layer compression, .PcRecord shows remarkable compression ratio in Fig 9. Additionally, our format standardization module supports the input of slice number, allowing the point cloud dataset to be converted into multiple .PcRecord files for storage, thus avoiding excessively large files. If adopted, the File Header will also include file path of each slice. 

To achieve faster loading performance, we also designed an efficient metadata management structure for .PcRecord, which can effectively improve the loading speed of point cloud. MainStream Deep learning frameworks mentioned before all process point cloud datasets as individual samples, employing on-demand loading to read data into memory. This approach typically uses a fixed loading logic, where all sample indices are stored and file paths are associated with the corresponding indices. Data is then loaded from its path from local disk when required. However, during this process, different deep learning frameworks employ varying metadata management strategies, and the architectures of dataset generators also differ, which leads to varying amounts of time spent on reading and preprocessing process during I/O operations.

The metadata of a point cloud sample in .PcRecord format relies on the following data structure:

\begin{lstlisting}
Metadata = std::tuple<TaskType, 
           //TaskType
           std::tuple<int, int>,   
           //Data Mapping
           std::vector<uint64_t>,
           //Sample Metadata
           json>;                
           //Scalar Metadata
\end{lstlisting}

\textbf{TaskType} indicates whether the task is a common task (kCommonTask) or a padded task (kPaddedTask), such as virtual tasks added for distributed training.

\textbf{Data Mapping} includes Shard IDs and Group IDs, which store the grouping levels of the samples. These mappings allow efficient division of the dataset into manageable chunks for parallel processing

\textbf{Sample Metadata} is represented as a vector of length 2 ([blob\_start, blob\_end]), specifying the precise byte range for the sample. It is the most critical part of the metadata, enabling efficient data access by providing accurate byte offsets for individual samples.

\textbf{Scalar Metadata} contains key-value pairs for storing additional metadata specific to the dataset, such as dataset-specific attributes, annotations, or configuration parameters.

Program caches all metadata related to tasks and samples, organizing task information (TaskType) and sample information (Data Mapping, Sample Metadata, Scalar Metadata) into task list and sample meta list, respectively. This design significantly enhances access and query efficiency. By preloading detailed metadata into memory, the program can quickly retrieve specific sample details and execute high-speed I/O operations. When accessing a sample, the program identifies and opens the corresponding shard file based on the shard\_id and group\_id, locates the appropriate group, and uses the [blob\_start, blob\_end] byte range to extract the sample. Finally, the sample's features are parsed using the JSON scalar metadata. This streamlined process ensures efficient and precise data access. Table 2 and Table 3 specifically demonstrates the high-speed loading performance enabled by this metadata management method.

\subsection{Distributed processing module}
To hand the large-scale point cloud datasets, which can easily reach hundreds of gigabytes, we introduce data parallelism strategy. More specifically, our system combines OBS (Object-Based Storage) streaming dataset loading method with data parallelism to accelerate and facilitate the loading and processing of point cloud datasets. The module's specific workflow is shown in Fig 5.
\begin{figure}[!h]
    \centering
    \includegraphics[width=1\linewidth]{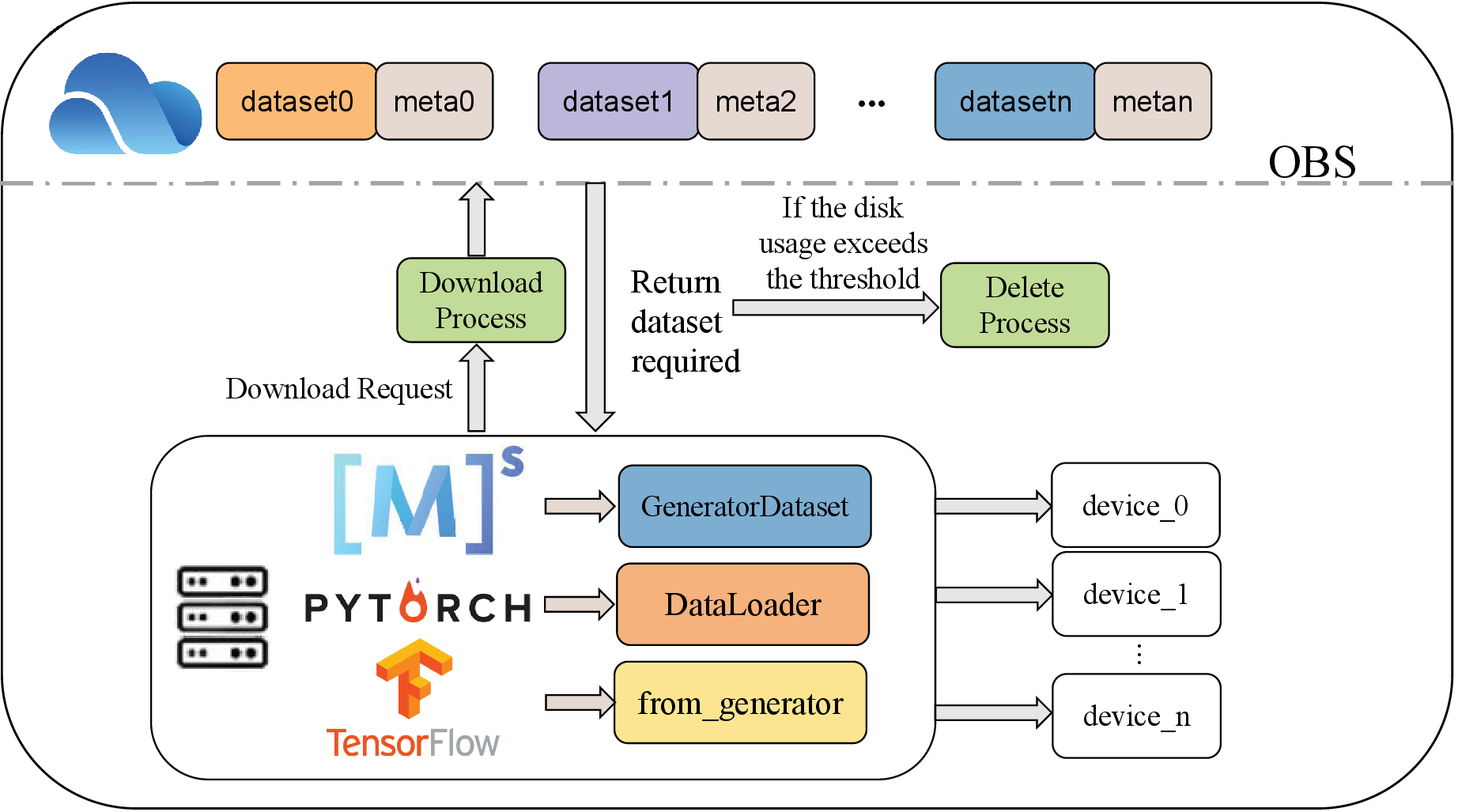}
    \caption{Workflow of OBS streaming dataset loading system.}
    \label{fig:enter-label}
\end{figure}

From the user perspective, the point cloud dataset is loaded with a main process, a download subprocess and a delete subprocess. The main process on server system is responsible for 1.the streaming loading of the point cloud dataset,2. point cloud data processing, 3. sending the processed data to the device. Streaming loading is achieved using a two-tier pipeline and the system loads dataset files based on sample index. The download subprocess is responsible for downloading specified dataset files from OBS and ensuring their integrity. And the delete subprocess monitors disk usage during training and deletes used dataset files when disk space exceeds a threshold.

The specific workflow of the module can be described as follows. First, main process downloads all meta index information of the point cloud dataset from OBS. Then system detects the dataset required by device0 for the current epoch as well as its corresponding index range required by device0. After that, the main process generates a download request for required files and sends it to the download process. When all files needed for the current epoch are downloaded for device0, the data is distributed across multiple devices.
\begin{figure}
    \centering
    \includegraphics[width=1\linewidth]{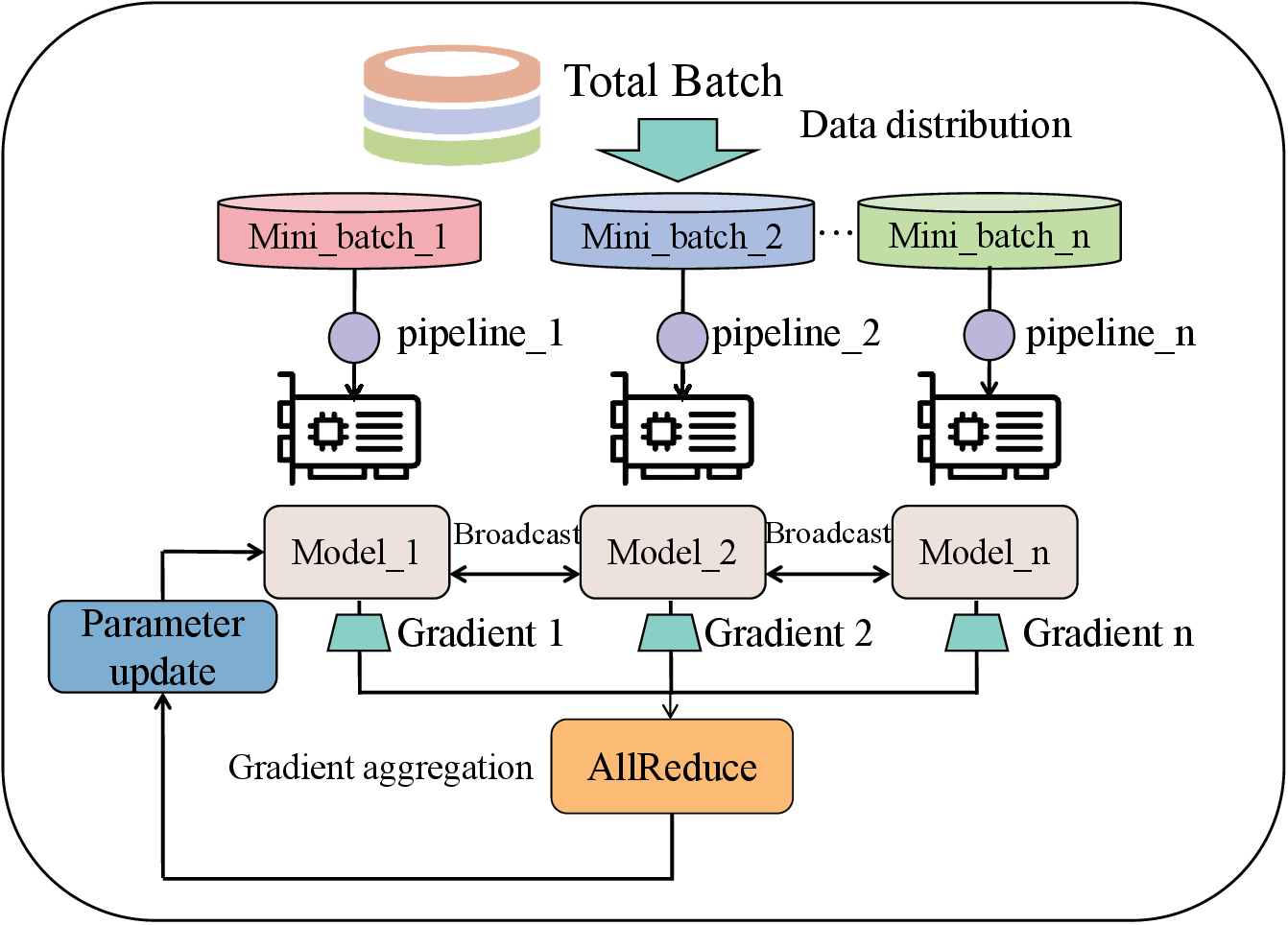}
    \caption{Workflow of data parallel processing and training on multiple devices.}
    \label{fig:enter-label}
\end{figure}

The distribution of dataset also plays an important role in this module. When all dataset is ready, the complete dataset (total batch) is divided into multiple subsets (mini-batches), with each subset assigned to a different process (in our experiments, we used 8 GPUs or NPUs, so there are 8 processes in total). The specific process is shown in Fig 6 and Algorithm 1. Each process runs an independent data processing pipeline, processing each mini-batch with identical hyperparameters. The initialization values of model parameters must be consistent. The network then performs forward and backward propagation on each mini-batch and gets the gradient of each training process. After that, an AllReduce operator is inserted to aggregate gradients across devices, followed by model parameter updates. This design improves data processing efficiency when handling point cloud data while ensuring the stability and consistency of the training process.

\begin{algorithm}[!h]
\caption{Data Parallel Training Workflow Algorithm}
\KwIn{total\_batch, num\_devices, num\_epochs}
\KwOut{Updated model parameters}

\textbf{1. Initialize devices and distribute the total batch into mini-batches}\;
num\_devices $\gets n$ \tcp*[f]{Number of devices}\;
mini\_batches $\gets$ split(total\_batch, num\_devices) \tcp*[f]{Split total batch into mini-batches}\;

\textbf{2. Initialize models and gradients for each device}\;
models $\gets$ [init\_model() for $i$ in $0 \dots$ (num\_devices $- 1$)]\;
gradients $\gets$ [None for $i$ in $0 \dots$ (num\_devices $- 1$)]\;

\textbf{3. Broadcast model parameters to all devices}\;
\For{$i \gets 1$ \textbf{to} (num\_devices $- 1$)}{
    models[i].parameters $\gets$ models[0].parameters \tcp*[f]{Sync parameters}\;
}

\textbf{4-6. Iterate through epochs to train the model}\;
\For{$epoch \gets 1$ \textbf{to} num\_epochs}{
    \For{$device\_id \gets 0$ \textbf{to} (num\_devices $- 1$)}{
        gradients[device\_id] $\gets$ compute\_gradient(models[device\_id], mini\_batches[device\_id]) 
        
\tcp{Step 4: Compute gradients;}
    }

    aggregated\_gradient $\gets$ allreduce(gradients) 
    
    \tcp{Step 5: Aggregate gradients;}

    \For{$model$ \textbf{in} models}{
        update\_parameters(model, aggregated\_gradient) 
        
        \tcp{Step 6: Update parameters;}
    }
}

\end{algorithm}

\subsection{Data loading and processing pipeline}
The data processing pipeline we designed is the core of the entire high-performance point cloud processing approach. Our approach integrates data processing deeply with Ascend Tensor Data Transmission and GPU Queue for network computation, achieving an end-to-end efficient point cloud network training process.

From the user perspective, once the point cloud dataset is loaded, users process the point cloud data with operators through API calls such as normalize. From the perspective of underlying code logic, the generation and execution of the data graph (which isn't discussed here) are actually responsible for the specific processing of point clouds. The pipeline generates data graph that organizes each data processing operation as nodes and uses optimizers for operator fusion and parameter adjustments to achieve a better efficiency.

The main algorithm of our pipeline is shown in Algorithm 2. We adopt a multi-stage parallel pipeline as the data processing pipeline. Operations such as normalization and translation (like Op a\_n and Op b\_n in Fig 1) are driven by a task scheduling mechanism that ensures each operator runs independently. All the map operators support multithread. The pipeline provides interfaces for adjusting the number of operation threads, allows flexible control the processing speed at each node.Since operators run independently, each operator’s output need connected to a buffer to place processed data. Therefore, We introduce the Connector, which is composed of a set of blocking queues, whose size dependent on the number of upstream and downstream thread, and two counters(pop\_from and expect\_consumer). Users can adjust the size of the Connector which allows the finer detail of computational resources. Back to Fig 1, during the flow of data, the Translate operator retrieves cached data of the upstream Normalize operator from Connector for processing.

\begin{algorithm}[!h]
\caption{Implementation Algorithm of Multi-Stage Parallel Pipeline}
\KwIn{producer\_size, consumer\_size}

\KwOut{Processed data in order}

\textbf{1. Initialize queues and counters}\;
queues 
$\gets [\text{empty queue}_1, \dots, \text{empty queue}_{producer\_size}]$ 

local\_queues 
$\gets [\text{empty queue}_1, \dots, \text{empty queue}_{consumer\_size}]$ 

expect\_consumer $\gets 0$ 

pop\_from $\gets 0$ 

\textbf{2. Implementation of key functions between unstream operator and Connector}\;

\SetKwFunction{FPush}{Push}
\SetKwProg{Fn}{Function}{:}{}
\Fn{\FPush{$worker\_id, data$}}{
    \uIf{$queues[worker\_id]$ is full}{
        \texttt{Wait\_to\_Push()}
        
    }
    \Else{
        $queues[worker\_id].$\texttt{add}($data$) 
        
        \Return \textbf{True} ;
    }
}

\SetKwFunction{FPop}{Pop}
\Fn{\FPop{$pop\_from$}}{
    \uIf{$queues[pop\_from]$ is empty}{
        \texttt{wait\_for\_data()}
    }
    \Else{
        $data \gets queues[pop\_from].$\texttt{pop}() 
        
        $pop\_from \gets (pop\_from + 1) \ mod \ producer\_size$ 

        \Return $data$ ;
    }
}

\textbf{3. Order-preserving strategy of the pipeline}\;
\tcp{Dequeue means pop.;}
\tcp{Enqueue means push;}

\While{any queue in $queues$ is not empty}{
    \If{queues[pop\_from] is not empty}{
        $data \gets dequeue(queues[pop\_from]) $
        
        $enqueue(data, local\_queues[expect\_consumer])$ 
        
        $pop\_from \gets (pop\_from + 1) \bmod \text{producer\_size}$
        
        $expect\_consumer \gets (expect\_consumer + 1) \bmod \text{consumer\_size}$

    }
    
    \If{expect\_consumer = 0}{
        \textbf{4. Process all consumer queues in current batch}\;
        \For{$j \gets 1$ \textbf{to} conumer\_size}{
            \While{local\_queues[$j$] is not empty}{
                $processed\_data \gets dequeue(local\_queues[$j$]) $

            }
        }
    }
}
\end{algorithm}

In this data processing mechanism, the order-preserving handling of point cloud data is crucial for ensuring training accuracy\cite{akbari2022deep}. The data processing pipeline provides an order-preserving mechanism for point cloud data. It uses a polling algorithm to ensure the orderliness of data during multi-threaded processing. The implementation details are shown in Fig 7.
\begin{figure}
    \centering
    \includegraphics[width=1\linewidth]{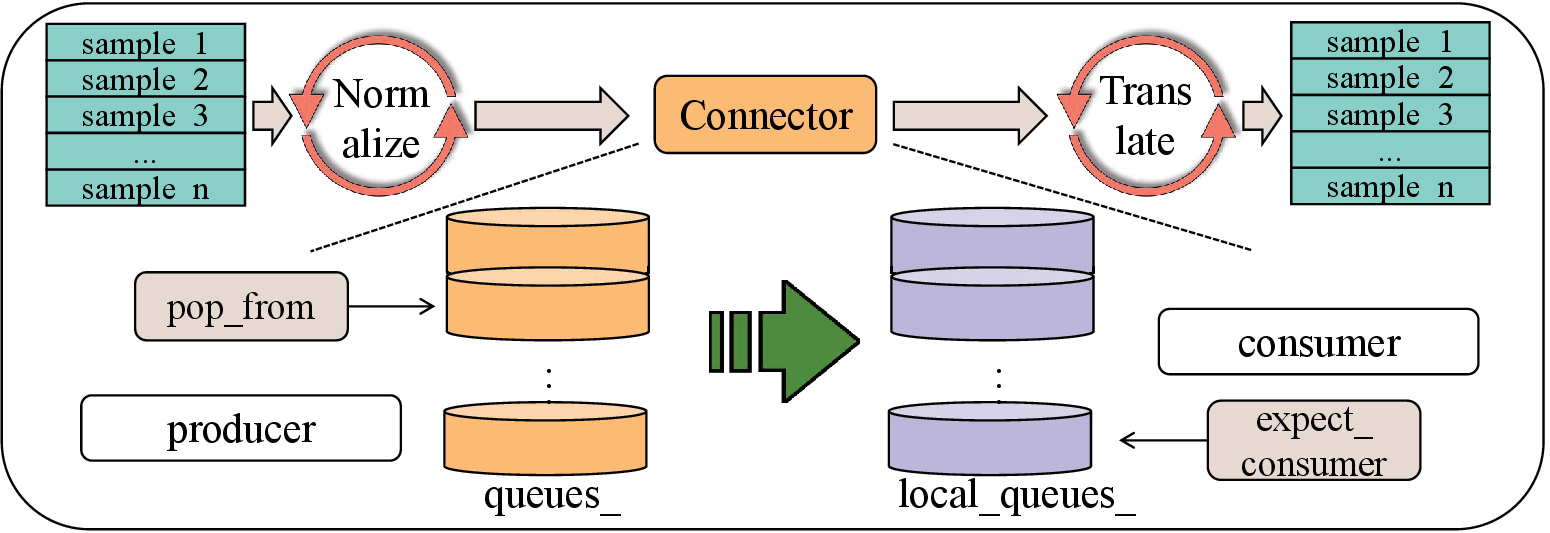}
    \caption{Diagram of order-preserving strategy of data processing pipeline.}
    \label{fig:enter-label}
\end{figure}
In the two operations shown Fig 7, the order-preserving operation occurs when the downstream Translate operation ask for data, where data is retrieved in a single-threaded polling manner from the upstream queue. The Connector has two counters: \verb|expect_consumer|, which tracks the number of consumers that have extracted data from the queues, and \verb|pop_from|, which indicates the internal blocking queue from which the next extraction will take place. \verb|expect_consumer| is managed through modulo operations on the number of consumers, while \verb|pop_from| performs corresponding operations for the producers. When \verb|expect_consumer| reaches zero again, it indicates that all the \verb|local_queues_| have completed processing the previous batch, allowing the system to continue assigning and processing the next batch, thus enabling multi-concurrent, order-preserving processing from upstream to downstream map operations.

This mechanism builds an end-to-end pipeline for data processing and network computation. Data processing keeps running, and the processed data is sent to the device-side cache. At the end of a training step, the network directly reads the next step's data from the device-side cache, forming a complete training process pipeline. Therefore, as long as the queue between device and network is not empty, model training will not block due to waiting for data.  
+
\subsection{Autotune module}
Due to the typically large size of point cloud datasets and the high dimensionality of the coordinates, data processing operations like normalization (which may include centroid subtraction, division by maximum distance, etc.), translation, and others generate a large number of intermediate results. These results consume significant pipeline resources. To maximize the efficiency of our point cloud data processing pipeline and avoid I/O bottlenecks, we introduce an autotune strategy. This strategy can automatically adjust the parallelism of the pipeline by detecting hardware and pipeline information, thereby maximizing system resource utilization to accelerate the data processing pipeline.

Throughout the whole loading process, the autotune strategy module continuously monitors whether the performance bottleneck lies on the data side or the network side. If the bottleneck is detected on the data side, the module will further adjust various operations within the data processing pipeline by tuning parameters. 

The workflow of the autotune strategy module consists of the following steps. First, it samples and gathers resource statistics from the data processing pipeline at user-defined intervals. Once sufficient information is collected, the module analyzes whether the performance bottleneck is on the data side. If so, an independent monitoring thread is launched to collect data processing delays, operator output queue utilization, CPU utilization, and other relevant metrics.

\begin{algorithm}[!h]
\caption{Autotune Optimize Algorithm}
\KwIn{Dataset $D$}
\KwOut{Best optimize result $R^*$}

\textbf{1. Initialize search parameters and variables}\;
Search\_Params $S \gets \{\}$
Number\_of\_Iterations $n \gets 10$ \;
Initial\_Points $I \gets \{\}$ \;
Save\_Path $P \gets "./"$ 

\textbf{2. Get current hyperparameters}\;
$H \gets \texttt{GET\_CURRENT\_HYPERPARAMS}(D)$\;

\textbf{3. Generate random search space}\;
$S \gets \texttt{RANDOM\_SEARCH}(D, S)$\;

\textbf{4. Define optimizer with Bayesian optimization}\;
optimizer $\gets \texttt{BAYES\_OPTIMIZATION}(f = \texttt{pipeline\_optimize}, \texttt{pbounds}=S, \texttt{rate}=H)$\;

\textbf{5. Perform optimization}\;
optimizer.\texttt{min}(\texttt{init\_points} = $I$, \texttt{n\_iter} = $n$)\;
$R^* \gets \texttt{optimizer.result}$\;
\texttt{SAVE\_PARAMS}($R^*$, $P$)\;
\Return $R^*$
\end{algorithm}

Next, the program uses the collected data to call relevant APIs to retrieve operator information. Based on this information, an initial analysis is performed, and hyperparameter search begins. 

Since the data augmentation operations on point cloud data are diverse, the search space is vast. To reduce the search space, the program first employs random search and then iteratively refines the search using Bayesian optimization\cite{bergstra2012random,claesen2015hyperparameter}. By combining random search and Bayesian optimization techniques, the hyperparameters (parallelism level and thread size for each data processing operation) obtained from each iteration are used to invoke the operators and generate new configuration data for the next iteration\cite{gulcu2019survey}. This iterative process continues until the optimal hyperparameters are found. The program saves the best hyperparameter results in a JSON file format first. On the next pipeline startup, the optimized hyperparameter set can be retrieved by deserializing the JSON file, ensuring that the data processing pipeline starts with high performance from the beginning of model training. The whole process can be described as Algorithm 3.
\begin{figure}[!h]
    \centering
    \includegraphics[width=1\linewidth]{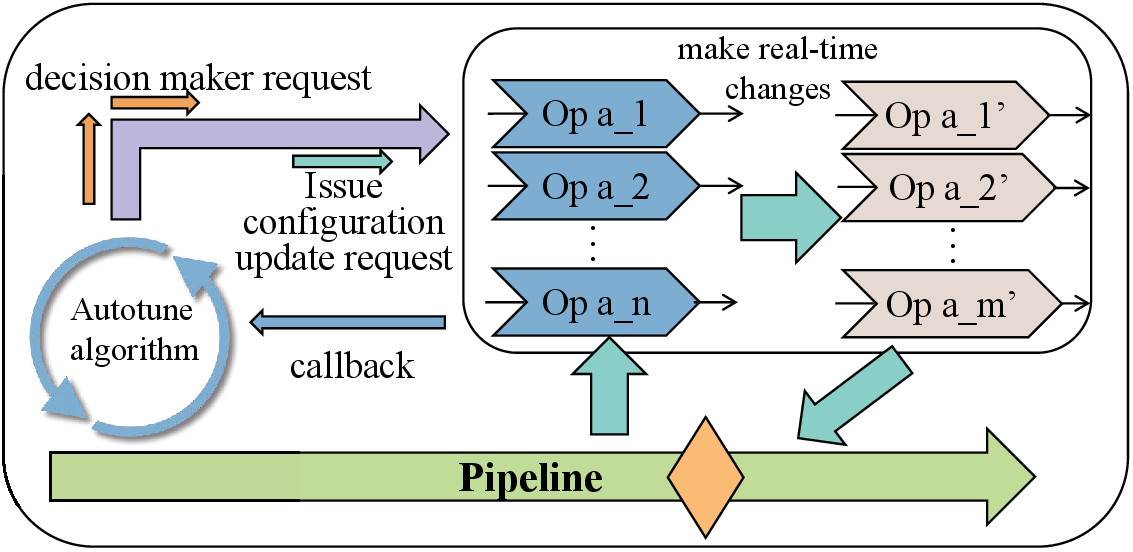}
    \caption{Workflow of pipeline update process.}
    \label{fig:enter-label}
\end{figure}

The next step of the autotune strategy module is to modify the dataset pipeline according to the optimal hyperparameters. The process is shown in Fig 8. During pipeline initialization, a real-time monitoring thread collects relevant information. Based on the configuration updates requested by the decision-maker in autotune module, the update requests for different operators are executed sequentially. Every so often, operators provide feedback through a callback to receive the next modification request, ensuring thread safety. The primary parameters that can currently be adjusted are as follows:

\begin{enumerate}
    \item \textbf{Adjusting the number of parallel threads}: This aims to improve the processing efficiency of the algorithm. When resources are limited, excessively increasing the number of parallel threads for operators may lead to CPU resource contention, which affects overall processing performance and reduces efficiency.
    \item \textbf{Optimizing the output queue size}: To improve operator processing efficiency and expand the capacity of each operator's output queue, the queue size should be smaller than the processing capability of the preceding operator. If the queue size continues to grow, it may lead to excessive memory consumption and trigger an out-of-memory (OOM) error when resources are insufficient.
    \item \textbf{Merging and Splitting in Map operators}: This strategy is used to balance the load on operator threads. By replacing common operators with more efficient implementations or merging multiple operator operations, thread overhead can be reduced, significantly improving overall data processing efficiency.

\end{enumerate}

\begin{table}[htbp]
\caption{The processing configurations of different point cloud datasets.}
\label{tab:dataset_configs}
\centering
\renewcommand{\arraystretch}{1.2} 
\setlength{\tabcolsep}{2pt}       
\footnotesize 
\begin{tabular}{@{}>{\centering\arraybackslash}p{1.5cm}>{\centering\arraybackslash}p{1cm}>{\centering\arraybackslash}p{4.4cm}>{\centering\arraybackslash}p{1.2cm}@{}}
\toprule
\textbf{Dataset}  & \textbf{workers} & \textbf{map operator}                                              & \textbf{batch size} \\
\midrule
ModelNet40         & 8                               & normalize, translate                                               & 32                  \\
S3DIS              & 8                               & normalize, jitter, rotate, translate                               & 4                   \\
ShapeNet           & 8                               & random scale, translate                                             & 10                  \\
Kitti              & 8                               & normalize                                                          & 32                  \\
Sunrgbd            & 8                               & Augment RGB color, random scale, rotate, flip along the YZ plane   & 4                   \\
ScanNet            & 8                               & Random Cropping, Downsampling                                      & 32                  \\
\bottomrule
\end{tabular}
\end{table}

\subsection{Comparative analysis of framworks' performance}
In section 4.3, we compared the performance of different deep learning frameworks (MindSpore, PyTorch, TensorFlow, Keras) in terms of the time required to load large-scale point cloud datasets. While a performance comparison was provided, there is still room for deeper insights into why certain frameworks perform better under specific conditions. The differences in loading times can be attributed to several factors:
Memory Management:
Each framework employs distinct memory management strategies that can influence data loading performance. For instance, TensorFlow and PyTorch both utilize internal memory pools to optimize memory allocation. PyTorch, with its dynamic computation graph, may better leverage memory for certain types of data. In contrast, MindSpore has its own memory allocation and release strategies tailored for hardware acceleration, which may lead to better memory utilization in specific conditions. The differences in how these frameworks allocate and release memory can lead to varied loading times when processing the same point cloud data.

\section{Experiments}
\renewcommand{\arraystretch}{1.5} 

\setlength{\tabcolsep}{2pt}  

\begin{table*}[htbp]
\centering
\footnotesize
\caption{Performance comparison of various frameworks on different datasets with GPU.The three values for each dataset are Cost Time (s), Average Memory Usage (\%), and Average CPU Usage (\%).The best values are shown in bold.}
\begin{tabular}{|l|c|c|c|c|c|c|c|c|c|c|c|c|c|c|c|c|c|c|}
\hline
           Dataset& \multicolumn{3}{|c|}{ModelNet40} & \multicolumn{3}{|c|}{S3DIS} & \multicolumn{3}{|c|}{ShapeNet} & \multicolumn{3}{|c|}{Kitti} & \multicolumn{3}{|c|}{Sunrgbd} & \multicolumn{3}{|c|}{ScanNet} \\ \hline 
 Metric& \multicolumn{18}{|c|}{Cost Time (s), Average Memory Usage (\%), Average CPU Usage (\%)}\\
\hline
MindSpore  & 17.24& 0.45\%   & 93.37\%   & 4     & 1.24\% & 128.10\% & 68.15  & 0.55\%  & 94.71\%   & 8.94  & 0.45\% & 103.51\% & 25.85  & 2.69\%  & 48.18\%  & 6.8    & 0.72\%  & 97.81\%  \\
\hline
PyTorch    & 39.83   & 0.37\%   & 21.55\%   & 6.26  & \textbf{1.13}\% & 99.38\%  & 7.99   & 0.32\%  & 66.40\%   & 9.18  & \textbf{0.32}\% & 54.44\%  & 18.57  & 2.84\%  & 61.50\%  & 7.06   & 0.59\%  & 90.03\%  \\
\hline
Tensorflow & 76.05   & 1.57\%   & 98.72\%   & 16.29 & 1.74\% & 106.81\% & 59.27  & 1.54\%  & 103.47\%  & 19.4  & 0.49\% & 99.60\%  & 29.27  & 3.93\%  & 91.43\%  & 16.96  & 1.00\%  & 70.92\%  \\
\hline
Keras      & 85.29   & 0.39\%   & 49.10\%   & 18.34 & 1.06\% & 54.76\%  & 63.69  & 0.38\%  & 51.73\%   & 35.06 & 0.51\% & 52.78\%  & 41.97  & 2.47\%  & 34.87\%  & 36.71  & 0.70\%  & 45.86\% \\
\hline
 Ours       & \textbf{2.61}    & \textbf{0.24}\%   & \textbf{114.26}\%  & \textbf{2.33}  & 1.24\% & \textbf{128.41}\% & \textbf{3.58}   & \textbf{0.27}\%  & \textbf{121.47}\%  & \textbf{2.89}  & 0.43\% & \textbf{169.15}\% & \textbf{2.3}    & \textbf{0.27}\%  & \textbf{115.90}\% & \textbf{1.2}    & \textbf{0.26}\%  &\textbf{108.12}\% \\
\hline
\end{tabular}
\end{table*}

\begin{table*}[!h]
\centering
\footnotesize
\caption{Performance comparison of various frameworks on different datasets with Ascend.The three values for each dataset are Cost Time (s), Average Memory Usage (\%), and Average CPU Usage (\%).The best values are shown in bold.}
\begin{tabular}{|l|c|c|c|c|c|c|c|c|c|c|c|c|c|c|c|c|c|c|}
\hline
           Dataset
& \multicolumn{3}{|c|}{ModelNet40} & \multicolumn{3}{|c|}{S3DIS} & \multicolumn{3}{|c|}{ShapeNet} & \multicolumn{3}{|c|}{Kitti} & \multicolumn{3}{|c|}{Sunrgbd} & \multicolumn{3}{|c|}{ScanNet} \\ \hline 
 Metric& \multicolumn{18}{|c|}{Cost Time (s), Average Memory Usage (\%), Average CPU Usage (\%)}\\
\hline
MindSpore  & 12.8    & 0.11\%   & 103.03\%  & 14.87 & 0.43\% & 105.67\% & 10.37  & 0.10\%  & 127.37\%  & 3.91  & 0.18\% & 109.50\% & 21.49  & 0.88\%  & 120.84\% & 3.12   & 0.22\%  & 99.21\%  \\
\hline
PyTorch    & 39.62   & 0.12\%   & 36.77\%   & 26.65 & 0.45\% & 95.26\%  & 5.1    & \textbf{0.09}\%  & 118.98\%  & 11.1  & \textbf{0.12}\% & 36.42\%  & 17.78  & 0.88\%  & 98.78\%  & 2.51   & 0.20\%  & 99.62\%  \\
\hline
Tensorflow & 77.76   & 0.14\%   & 107.27\%  & 34.46 & 0.47\% & 117.58\% & 35.65  & 0.17\%  & 115.57\%  & 24.26 & 0.20\% & 104.94\% & 24.46  & 0.93\%  & 120.72\% & 11.17  & 0.26\%  & 109.10\% \\
\hline
Keras      & 77.94   & 0.12\%   & 106.92\%  & 36.65 & 0.33\% & 111.53\% & 35.34  & 0.12\%  & 113.82\%  & 23.51 & 0.20\% & 105.40\% & 25.05  & 0.86\%  & 118.05\% & 12.9   & 0.26\%  & 108.86\% \\
\hline
 Ours       & \textbf{1.84}    & \textbf{0.08}\%   & \textbf{161.00}\%  & \textbf{7.92}  & \textbf{0.12}\% & \textbf{283.55}\% & \textbf{3.95}   & \textbf{0.09}\%  & \textbf{150.58}\%  & \textbf{1.71}  & 0.19\% & \textbf{299.73}\% & \textbf{} \textbf{0.7}    & \textbf{0.08}\%  & \textbf{164.66}\% & \textbf{0.13}   & \textbf{0.07}\%  &\textbf{189.77}\% \\
\hline
\end{tabular}
\end{table*}

In the experiment, to thoroughly demonstrate the superiority of our proposed data processing pipeline, we selected six of the most popular point cloud datasets. These datasets cover a range of different scenarios, such as autonomous driving, indoor scene modeling, and 3D object recognition. Each dataset has unique characteristics, including data scale, point cloud density, and scene complexity, which allows for a comprehensive assessment of the adaptability and performance of our system. Specifically, we conducted data loading and processing experiments on these datasets, with the processing methods outlined in Table 1.

To ensure fairness and comparability, we performed the same operations on different deep learning frameworks, ensuring that all comparative results were obtained under consistent conditions. It should be noted that the num\_parallel\_workers parameter in Table 1 represents the initial setting value; if the user activates the autotune module, this parameter may be dynamically adjusted during the program's execution.

During the experiment, we repeated tests multiple times and collected several key performance metrics. These metrics include the time consumption for the entire loading and processing workflow, average memory usage, and average CPU usage. We conducted experiments on two hardware platforms (GPU and Ascend), with detailed hardware and software configurations provided in section 4.1. The final experimental results are summarized in Tables 2 and 3, where the best results for each group of data are marked in bold. From Tables 2 and 3, it is evident that our data processing pipeline achieved optimal performance in nearly all experiments, showcasing a significant efficiency advantage.

To further validate the superiority of our system, we designed several extension experiments:

We visually compared the storage space consumption of the point cloud datasets after enabling the format unification module.
Through multiple rounds of comparative experiments, we verified the stability of the point cloud data loading and processing process across different frameworks.
To test the performance of our system in a distributed environment, we conducted experiments in a multi-device setup.
We compared the throughput performance of different frameworks when loading datasets.
We performed a detailed performance analysis on the Kitti dataset.

\renewcommand{\arraystretch}{1.5} 
\setlength{\tabcolsep}{2pt}

\begin{table}[htbp]
\caption{Experiment Setting}
\label{tab:experiment_setting}
\centering
\footnotesize
\renewcommand{\arraystretch}{1.0} 
\scriptsize 
\begin{tabular}{@{}p{3.2cm}p{4.6cm}@{}}
\toprule
\textbf{Experiment}        & \textbf{Categories} \\
\midrule
\textbf{Hardware Combination} & \textbf{\textit{NPU+CPU}}
\newline \textbf{\textit{GPU+CPU}}\\
\midrule
\textbf{Framework Version}             & TensorFlow 2.13.1,PyTorch 2.4.0,MindSpore 2.3.1,Keras 2.13.1\\
\bottomrule
\end{tabular}
\end{table}

\subsection{Experimental setup}
\textbf{Datasets and Deep Learning Frameworks:}
In our experiments, we load and process six most commonly used point cloud datasets, including Kitti (intercepting the first 100 .bin files), ModelNet40, ShapeNet, S3DIS , ScanNet, and SUN RGB-D. These datasets cover outdoor autonomous driving and indoor scenes, encompassing multiple 3D vision tasks such as 3D semantic segmentation, 3D object detection and 3D Shape Classificatio. The sizes of these datasets range from 3GB to 100GB.

Since experiments in our study were conducted on limited datasets, it is important to discuss how our system can be applied to other point cloud data formats that were not included in the study. As mentioned before, our format standardization module is tightly integrated with MindSpore, and this integration ensures that the system can be easily extended to different point cloud data formats.

In practical use, as long as MindSpore's built-in data generator (GeneratorDataset) is able to successfully generate the data from the original format loaded by data loader class user define, the dataset loaded can then be converted into the .PcRecord format. It means once the user defines a standard and reasonable data loader class that correctly interfaces with MindSpore, the dataset can be loaded into our system, and our metadata management system and efficient processing pipeline can be seamlessly applied, regardless of the original format. This makes our system flexible and scalable, with the ability to handle a wide variety of point cloud data formats.

To make this process clearer, we provide an using example of our system in the Appendix A, showing the format conversion process.


To prove superiority, We compared the performance of our point data processing pipeline against PyTorch, TensorFlow, Keras and the open-source framework MindSpore. PyTorch, TensorFlow, and MindSpore come with built-in data generator like \text{torch.utils.data.DataLoader}, \text{tf.data.Dataset.from\_generator} and \text{ds.GeneratorDataset} respectively. However, Keras requires users to implement their own data generator.

System Configuration: We conducted experiments on two devices.

\subsection{Point cloud data storage space comparison }

In the initial phase of the experiment, we applied a format unification operation to the raw point cloud data to ensure that the data could be loaded and processed more efficiently. As mentioned before, .PcRecord adpots a dual-layer mechanism to compress the size of the point cloud dataset. To demonstrate the efficiency of our compression mechanism, we conducted experiments on various point cloud datasets. We recorded the datasets' storage space before and after compression for the same sample size and calculated the compression ratio. Fig 9 shows a comparison of the storage space sizes of each point cloud dataset before and after the format conversion. (It should be noted that, for efficiency and consistency in measurement, only a subset of the datasets was used for the storage space evaluation.)
 \begin{figure}[!h]
     \centering
     \includegraphics[width=1\linewidth]{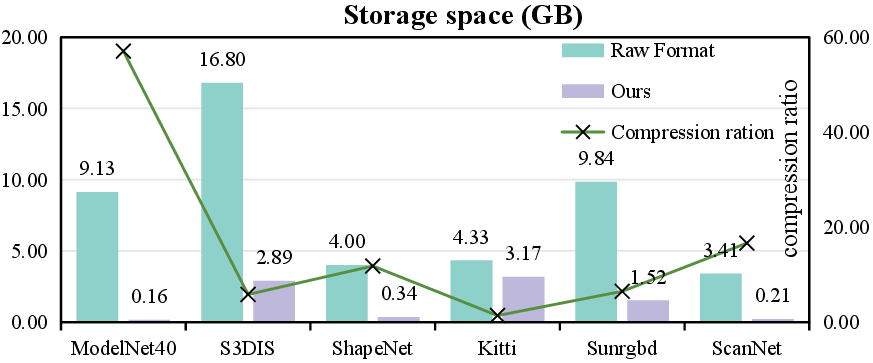}
     \caption{Comparison of storage space before and after the format conversion.}
     \label{fig:enter-label}
 \end{figure}
The experimental results show that, after the format unification process, the storage space of the different point cloud datasets was significantly compressed, with the following reductions: 57.06x (Kitti), 5.81x (ModelNet40), 11.81x (S3DIS), 1.37x (ShapeNet), 6.47x (SUN RGB-D), and 16.63x (ScanNet).

The compression effectiveness varied depending on the original format of the point cloud data. We observed that the Kitti dataset, which was originally in binary format, had a relatively high initial storage efficiency, resulting in a lower compression ratio of 1.37x. However, even so, our format unification operation still played an important role in optimizing the data structure and reducing file redundancy. In contrast, point cloud datasets stored in text formats, such as ModelNet40 and ShapeNet, experienced significantly higher compression, with reductions of 5.81x and 11.81x, respectively. This is because text formats inherently contain a large amount of extraneous characters, delimiters, and unnecessary whitespace. By converting to a unified, efficient binary format, these redundant data elements were completely eliminated, greatly reducing the storage requirements of the files.

\subsection{Time testing for single-device data loading and processing}
The core objective of this experiment is to evaluate the performance advantages of our proposed data processing engine in the point cloud data loading and processing processes. We focus on the most intuitive and critical metric—time consumption—by incorporating precise timing modules in the code to record the time taken by each framework when processing point cloud datasets. The experiment was rigorously tested on two hardware platforms, GPU and Ascend, using six point cloud datasets covering different application scenarios. It is important to note that, in the experiment, we ensured that the preprocessing process for each point cloud dataset was identical across all deep learning frameworks, and hyperparameters were strictly kept the same to ensure fairness and comparability of the results.

\begin{figure*}[!h]
     \centering
     \includegraphics[width=1\linewidth]{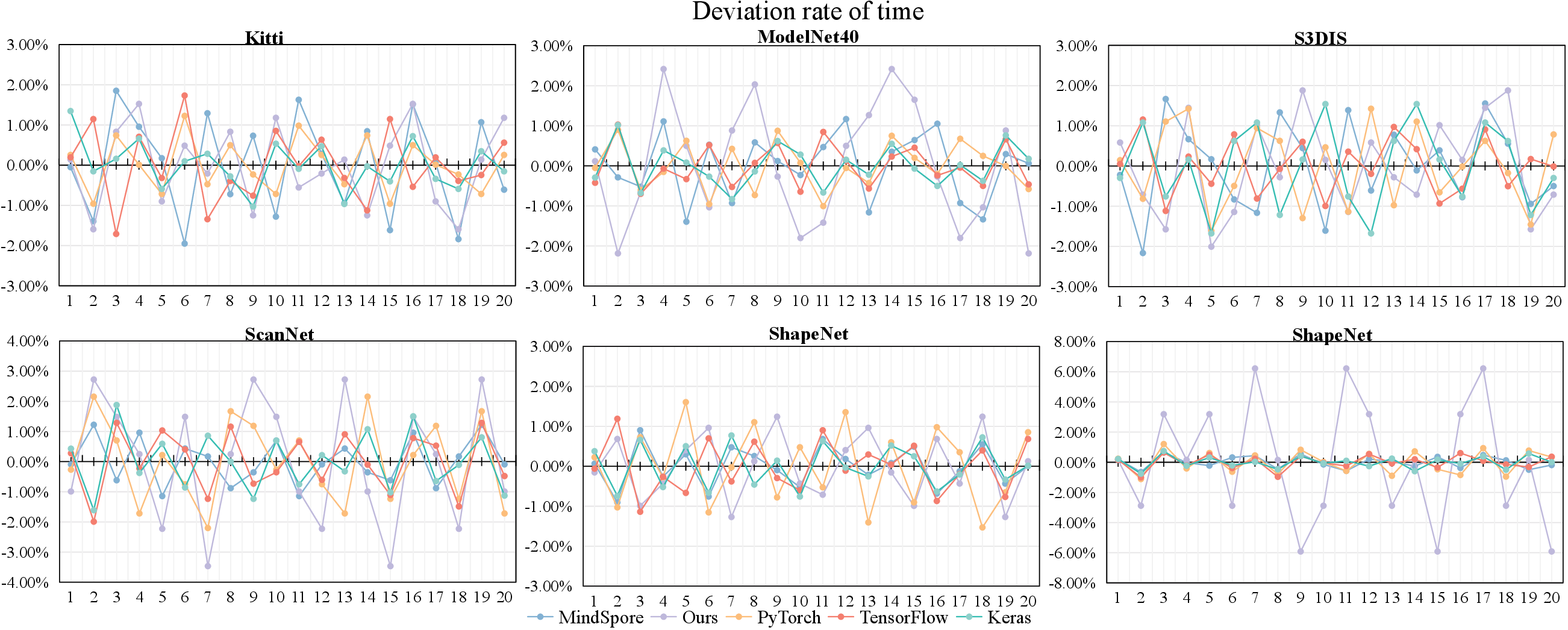}
     \caption{The deviation rate of time spent on data loading and processing with different datasets for 20 epochs.}
     \label{fig:enter-label}
\end{figure*}

To validate the reliability of the test results, we designed a 20-cycle experiment, recording the time consumption for each cycle and calculating the deviation rates. The experimental results are shown in Fig 10. Excluding one special case, all deviation rates were under 3\%, which fully demonstrates the rationality of the experimental setup and the reliability of the data collection process. The stable time consumption provides a solid foundation for the experimental results and ensures the validity of subsequent performance analysis.

We normalized the time consumption data obtained from the GPU and Ascend platform, and the results are presented in a bar chart in Fig 11. From the results, it is clear that PyTorch demonstrates a significant performance advantage over TensorFlow and Keras in terms of point cloud data loading and processing. The MindSpore framework achieves performance results comparable to or even better than PyTorch, especially exhibiting higher efficiency on certain datasets. Our data processing engine outperforms all other frameworks in terms of performance on every point cloud dataset. Specifically, compared to the optimal results of other frameworks, our system achieves performance improvements of 6.61x (ModelNet40), 2.69x (S3DIS), 2.23x (ShapeNet), 3.09x (Kitti), 8.07x (SUN RGB-D), and 5.67x (ScanNet) across the six datasets. These significant performance gains not only demonstrate the high efficiency of our system but also showcase its strong adaptability on GPU hardware platforms.

On the Ascend platform, we repeated the experiments with the exact same code, datasets, and hyperparameters. The results show that, similar to the GPU platform, our system still exhibits a strong performance advantage across all point cloud datasets. Compared to the optimal results of other frameworks, our system achieves performance improvements of 6.9x (ModelNet40), 1.88x (S3DIS), 1.29x (ShapeNet), 2.28x (Kitti), 25.4x (SUN RGB-D), and 19.3x (ScanNet). Notably, the performance improvement on the SUN RGB-D and ScanNet datasets on the Ascend platform is particularly remarkable, reaching 25.4x and 19.3x, respectively. This further proves that our system not only excels in GPU environments but also demonstrates excellent compatibility and performance advantages in NPU environments.

\begin{figure}[!h]
    \centering
    \includegraphics[width=1\linewidth]{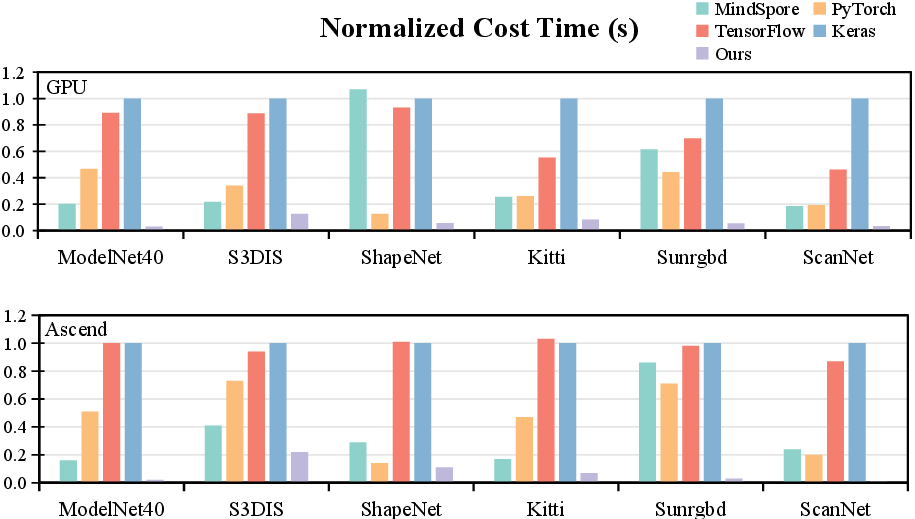}
    \caption{Normalized time consumption of loading and processing different datasets on GPU and Ascend.}
    \label{fig:enter-label}
\end{figure}

 \subsection{CPU usage comparison }
 In this experiment, we conducted a comprehensive comparison of the average CPU utilization across different frameworks while loading and processing six distinct point cloud datasets. Specifically, we tested MindSpore, PyTorch, TensorFlow, Keras, and our high-performance data processing engine, running experiments on both GPU and Ascend hardware platforms. During the tests, we monitored the CPU load in real-time for each data frame processed, calculating the average CPU utilization for each framework to assess the resource efficiency of different frameworks in the data processing pipeline.
\begin{figure}[!h]
    \centering
    \includegraphics[width=1\linewidth]{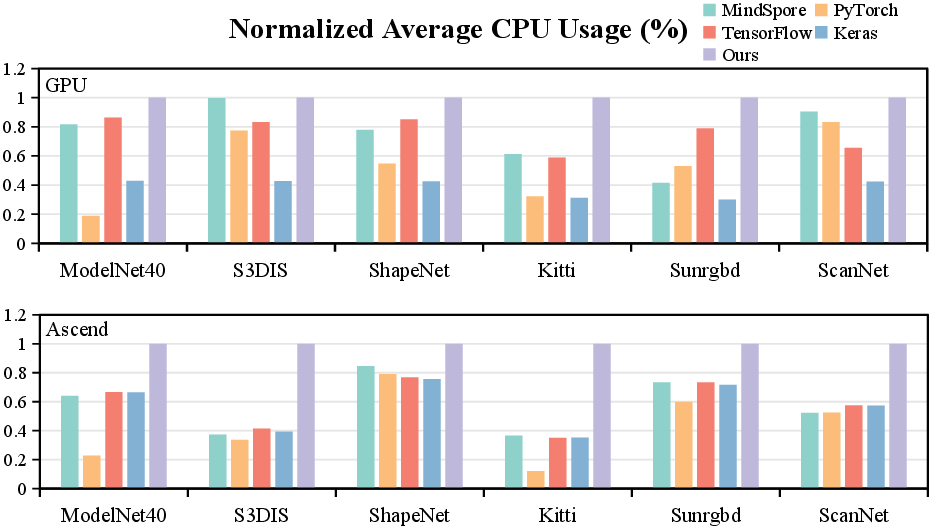}
    \caption{Normalized average CPU usage of loading and processing different datasets on GPU and Ascend.}
    \label{fig:enter-label}
\end{figure}
The normalized results are presented in Fig 12. Across all six point cloud datasets, our high-performance data processing engine consistently showed significantly higher average CPU utilization compared to other frameworks, demonstrating its outstanding resource utilization capabilities. This advantage is primarily attributed to the design characteristics of our data processing engine: an efficient data processing pipeline. Through deep optimization, this pipeline successfully achieves high parallelism in data loading and processing tasks. In contrast to traditional frameworks, our system effectively reduces I/O bottlenecks during data reading, enabling the CPU to fully leverage its computational potential.

Additionally, we observed that, whether on the GPU or Ascend platform, our system maintained a consistently high CPU utilization, indicating excellent adaptability and optimization across different hardware architectures. This cross-platform advantage not only enhances the overall efficiency of the data processing pipeline but also provides new insights for performance optimization in large-scale distributed training scenarios.

 \subsection{Distributed processing }
In this experiment, we employed distributed system to preprocess point cloud datasets by splitting the datasets and distributing them across eight different GPUs for parallel pipeline operations. It is important to note that since Keras does not provide an official data distribution interface, although users can implement distributed operations through custom methods, these approaches often significantly impact performance. Therefore, to ensure fairness and consistency in the experiment, we excluded distributed testing for Keras.

We performed performance tests on both GPU and Ascend platforms, evaluating the distributed processing of six point cloud datasets. During the experiment, we recorded the data loading and processing time for each framework in the distributed setting. To clearly present the performance comparison results, we normalized the test data and plotted the results as bar charts, as shown in Fig 13.

\begin{figure}[!h]
    \centering
    \includegraphics[width=1\linewidth]{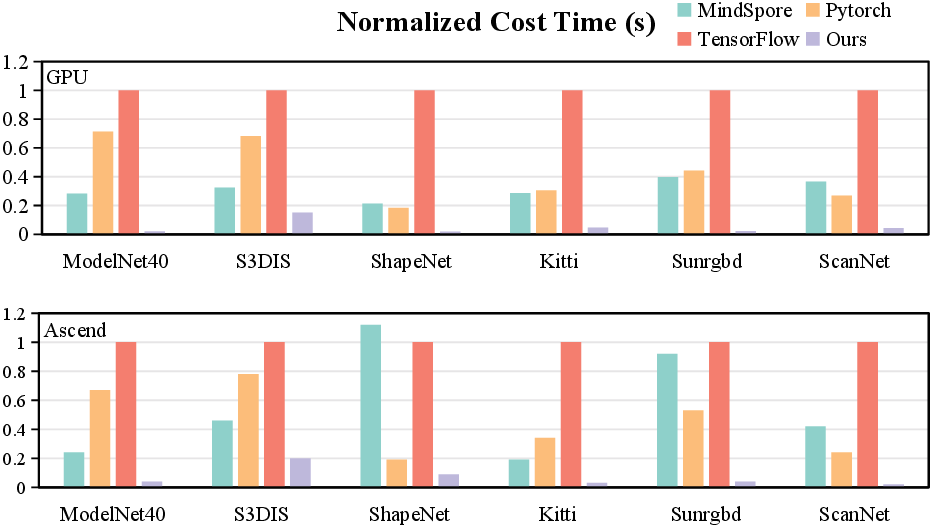}
    \caption{Normalized time consumption of reading and processing different datasets with the distributed setting.}
    \label{fig:enter-label}
\end{figure}

On the GPU platform, our high-performance data processing engine demonstrated significant advantages. Compared to the fastest distributed implementations in other frameworks, our system achieved performance improvements on six datasets by 38.05x (ModelNet40), 2.16x (S3DIS), 9.91x (ShapeNet), 6.12x (Kitti), 18.17x (Sun RGB-D), and 6.31x (ScanNet). Similarly, on the Ascend platform, our system still outperformed others, with speed improvements of 7.64x, 2.33x, 1.98x, 5.92x, 13.80x, and 15x on the aforementioned datasets. These results further confirm the strong adaptability and outstanding performance of our distributed data processing system across different hardware platforms.

\renewcommand{\arraystretch}{1} 
\setlength{\tabcolsep}{8pt}
\begin{table}[ht]
\centering
\small
\caption{Performance comparison between MindSpore and our method with single and multiple devices.}
\begin{tabular}{lcccc}
\toprule
\multirow{2}{*}{\textbf{Dataset}} & \multicolumn{2}{c}{\textbf{MindSpore}} & \multicolumn{2}{c}{\textbf{Ours}} \\ 
\cmidrule(lr){2-3} \cmidrule(lr){4-5}
                                  & \textbf{single}  & \textbf{mul}          & \textbf{single}  & \textbf{mul}       \\ 
\midrule
ModelNet40                        & 17.24            & 22.07                 & 2.61             & 0.58               \\ 
S3DIS                             & 4.00             & 5.94                  & 2.33             & 2.75               \\ 
ShapeNet                          & 68.15            & 13.57                 & 3.58             & 1.17               \\ 
Kitti                             & 8.94             & 6.73                  & 2.89             & 1.10               \\ 
Sunrgbd                           & 25.85            & 6.36                  & 2.30             & 0.35               \\ 
ScanNet                           & 6.80             & 2.50                  & 1.20             & 0.29               \\ 
\bottomrule
\end{tabular}

\label{tab:performance_comparison}
\end{table}

\begin{figure*}[!h]
    \centering
    \includegraphics[width=1\linewidth]{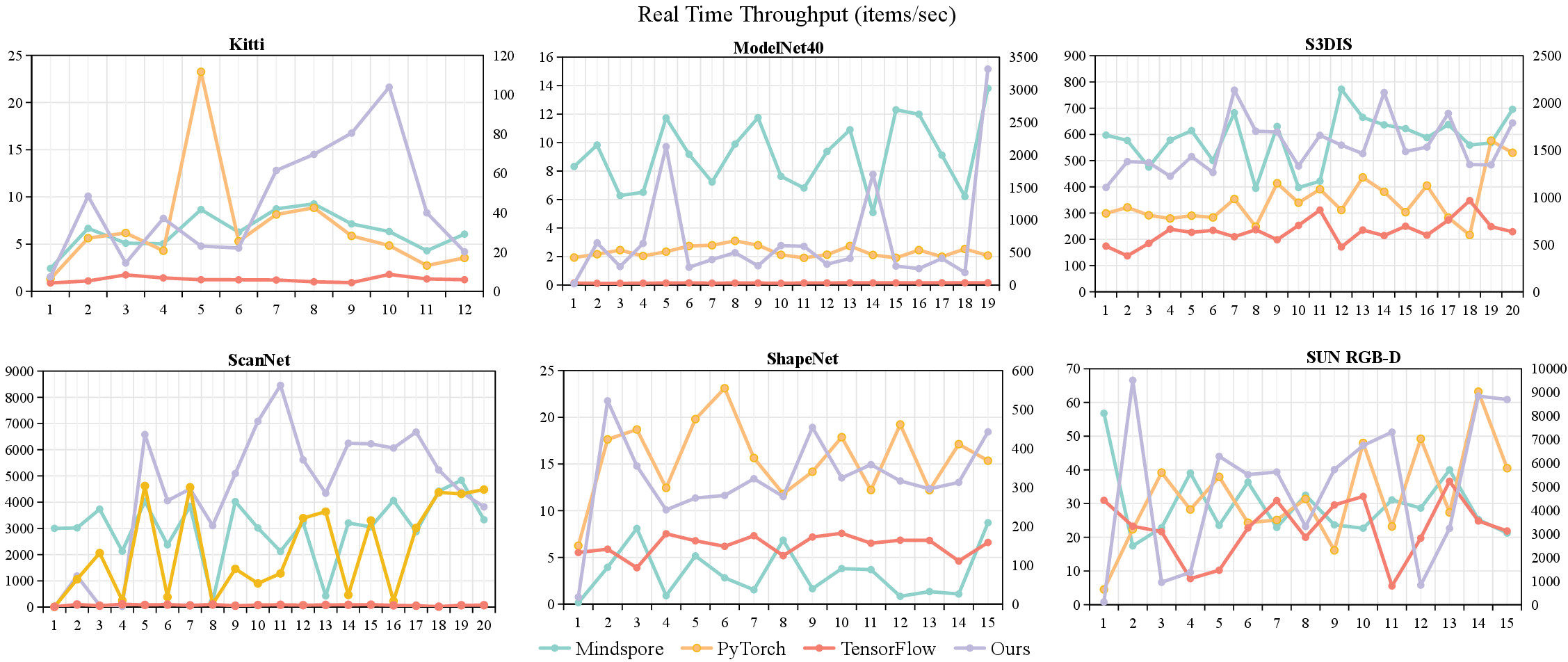}
    \caption{The real-time throughput(items/sec) when processing each batch of point cloud dataset.The right vertical axis is for our method if it exists.}
\end{figure*}
To further validate the effectiveness of data parallelism, we compared the performance of MindSpore and our system in both single-GPU and multi-GPU scenarios. In most experiments, multi-GPU operation significantly outperformed single-GPU performance. This improvement is primarily due to the multiple pipeline processes implemented across devices. These processes effectively distribute the data preprocessing tasks across different devices, alleviating the load on each device and preventing bottlenecks that may occur in a single-device setup.

\subsection{Throughput monitoring }

Throughput is a key metric for measuring the efficiency of data loading and processing, directly influencing the training speed and overall system performance. Throughput represents the number of samples that a data processing engine can load and process per unit of time. In the past, issues related to data loading in point cloud data processing were often tied to bottlenecks in throughput.

In this experiment, we primarily tested the throughput performance of deep learning frameworks while loading different datasets, focusing on each batch’s throughput. Each experiment was run with a fixed batch size. By monitoring the processing time for each batch when loading a dataset in each framework, we calculated the real-time throughput for each batch. The measured data are shown in Fig 14.

It is worth noting that an additional vertical axis is added to the right side of the line chart. This is because our method (Ours) achieved throughput far exceeding other frameworks on some datasets. To better display all results, we have created a separate vertical axis for "Ours." The experimental results indicate that our method demonstrates a significant advantage in throughput. Compared to the best results from other frameworks, our peak throughput improved by 4.46x, 240.19x, 2.76x, 1.74x, 22.59x, and 150.46x on the Kitti, ModelNet40, S3DIS, ScanNet, ShapeNet, and SUN RGB-D datasets, respectively. Additionally, for the lowest throughput (i.e., the worst-case performance), our method improved by 3.34x, 37.52x, 3.11x, 291x, and 48x on the Kitti, ModelNet40, S3DIS, ShapeNet, and SUN RGB-D datasets. These results not only show that our method significantly breaks through throughput limits but also highlight its ability to overcome performance bottlenecks under demanding conditions, greatly enhancing the stability of data loading and processing.

This significant improvement is due to key optimizations in our data processing engine. These improvements not only drastically reduced the processing time for each batch but also significantly increased the overall system’s resource utilization efficiency. In large-scale point cloud data training tasks, the increase in peak throughput can effectively shorten training cycles, while the improvement in lowest throughput ensures the reliability of long-running tasks, preventing system stagnation or resource wastage caused by performance bottlenecks.

\subsection{Real-time memory and CPU monitoring for the kitti dataset }
Autonomous driving datasets, such as Kitti, have more stringent real-time response and processing requirements compared to other datasets. When handling such data, efficient memory management and CPU resource utilization are critical. To validate the advantages of our system based on the MindSpore framework in terms of memory usage, we designed a real-time monitoring experiment to capture and analyze the memory consumption during the processing of the Kitti dataset with both our approach and the original MindSpore framework on Nvidia GPU device with 64-cores CPU.

For the experiment, we set up a monitoring thread to track memory usage and CPU utilization at intervals of 0.1ms. This allowed us to compare the performance of our system with the original MindSpore framework under the same task. In the experiment, we executed five data loading and processing cycles to simulate the data loading and processing loops typical in actual scenarios. 

The experimental results of memory usage are shown in Fig 15. During the five data loading and processing cycles, the memory usage of the original MindSpore framework exhibited a stepwise cumulative increase. After each data reading and processing task, memory consumption gradually increased, and this consumption did not decrease in subsequent cycles which  can been seen from an further experiment shown in Fig 16. This indicates that the original MindSpore framework has a memory management bottleneck when processing the KITTI dataset in a loop , failing to effectively release memory or balance memory load.

In contrast, the framework using our approach (Ours) showed significant advantages. We observed that the memory usage followed a "breathing" pattern, with distinct peaks and valleys. This pattern indicates that memory usage is dynamically adjusted through efficient memory recycling and redistribution mechanisms, effectively preventing resource blocking. Specifically, after each data processing cycle, memory is released in a timely manner, ensuring that no memory overflow or blocking issues occur in subsequent data processing.
 \begin{figure}[!h]
    \centering
    \includegraphics[width=1\linewidth]{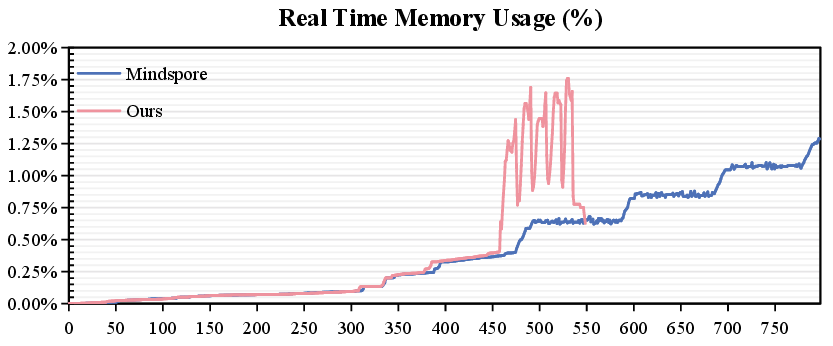}
    \caption{MindSpore vs. Ours in terms of CPU usage with Dataset Kitti.}
    \label{fig:enter-label}
\end{figure}
\begin{figure}[!h]
    \centering
    \includegraphics[width=1\linewidth]{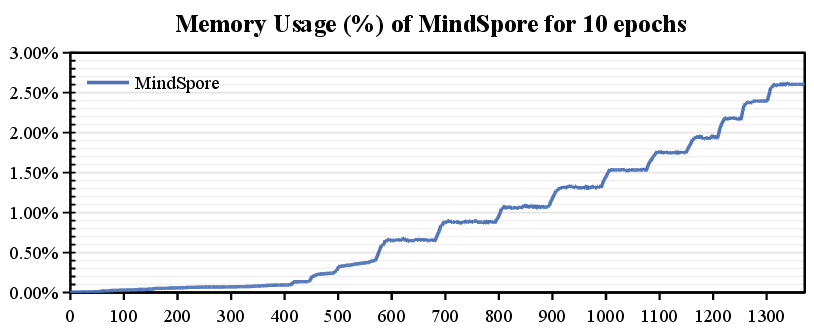}
    \caption{MindSpore's memory usage curve for 10 epochs with Dataset Kitti.}
    \label{fig:enter-label}
\end{figure}

In addition to memory usage, we also monitored CPU utilization. The results are shown in Fig 17. The smooth segment at the beginning represents the initialization phase of the system. We found that in the data processing cycle, the CPU utilization for the original MindSpore framework did not significantly change and remained at a relatively fixed level. In contrast, the CPU utilization for the framework using our system (Ours) significantly increased during the five Kitti dataset reading and processing cycles.

\begin{figure}[!h]
    \centering
    \includegraphics[width=1\linewidth]{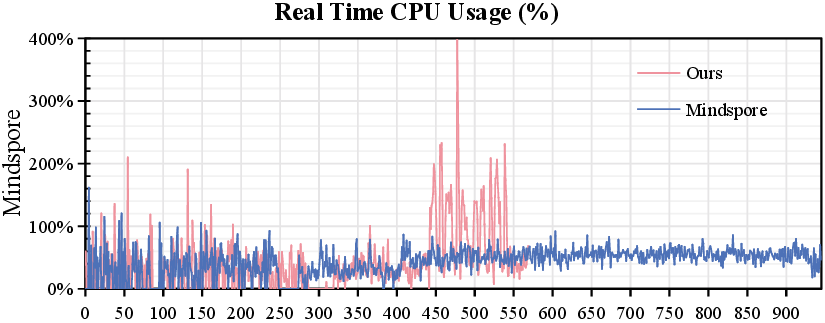}
    \caption{MindSpore vs. Ours in terms of memory usage with Dataset Kitti.}
    \label{fig:enter-label}
\end{figure}

This indicates that by adopting a multi-stage parallel pipeline design, our system can fully leverage the processing power of multi-core CPUs. During each data loading and processing cycle, multiple parallel tasks are effectively scheduled, and the CPU load is reasonably distributed across different operational stages. This optimization ensures that each task can make the best use of CPU resources, enhancing the overall data processing speed and ensuring that real-time processing requirements are met.

\section{Conclusion}
In this paper, we present a novel data reading and processing system for large-scale point cloud datasets. The experimental results show that, compared to mainstream frameworks such as PyTorch, TensorFlow, Keras, and MindSpore, this system significantly shortens the time taken to load point cloud datasets while making better use of computational resources.

The integration of the \textbf{.PcRecord} format further optimizes the complexity of format conversion and processing in point cloud datasets, thereby accelerating the reading process. The autotune strategy, with an independent monitoring thread, dynamically adjusts pipeline hyperparameters in real-time, minimizing I/O bottlenecks. The incorporation of distributed processing technology enables multi-card machines to further enhance efficiency. In order to solve the problem that point cloud data sets occupy too much storage space, we use OBS streaming data loading technology. Also, the data processing pipeline accelerates each dataset operation in parallel and strengthens the reliability of the pipeline by introducing an ordering mechanism.

This research highlights the potential of specialized data processing pipelines and formats in point cloud data processing, paving the way for more efficient and scalable deep learning workflows. Future work will explore more optimizations, including map operator optimization and adaptive learning techniques, to enhance deep learning models' capabilities in 3D vision tasks.

\appendix
\section{Code example of the format conversion process}

\begin{lstlisting}
class ModelNet40Dataset:
//The Data loader class user defines
    def __init__(self, root_path, split, use_norm, num_points):
        """read datasets path"""
        ......
    def __getitem__(self, index):
        """get item"""
        fn = self.datapath[index]
        label = self.classes[self.datapath[index][0]]
        label = np.asarray([label]).astype(np.int32)
        point_cloud = np.loadtxt(fn[1], delimiter=',').astype(np.float32)
        if self.use_norm:
            point_cloud = point_cloud[:self.num_points, :]
        else:
            point_cloud = point_cloud[:self.num_points, :3]
        point_cloud[:, :3] = self._pc_normalize(point_cloud[:, :3])
        return point_cloud, label[0]

    def translate_pointcloud(self,pointcloud):
        """translate"""
        ......

    def __len__(self):
        """len"""
        return len(self.datapath)
    
    def _pc_normalize(self, data):
    # get the centroid, and normalize to [0, 1]
        ......
\end{lstlisting}

\begin{lstlisting}
import mindspore
import mindspore.dataset as ds
import PcRecord
PcRecord_path = "/datasets/modelnet40.pcrecord"

dataset = ModelNet40Dataset(root_path=ModelNet40_dataset_dir,split='train',use_norm=True,num_points=1024)

dataset = ds.GeneratorDataset(source=dataset_generator,column_names=["data", "label"],shuffle=False,num_parallel_workers=16)

//Data generator
PcRecord.save(dataset,PcRecord_path)
//Convertion
dataset = PcRecord.load(dataset_files=PcRecord_path,["data", "label"])
PcRecord_path
//Load ModelNet40 in .PcRecord
dataset = dataset.map(operations=Modelnet40_normalize,num_parallel_workers=num_parallel_workers)
dataset = dataset.batch(batch_size,num_parallel_workers=num_parallel_workers)



\end{lstlisting}

\section{Application Scenarios}

\begin{figure}[!htbp]
    \centering
    \includegraphics[width=1\linewidth]{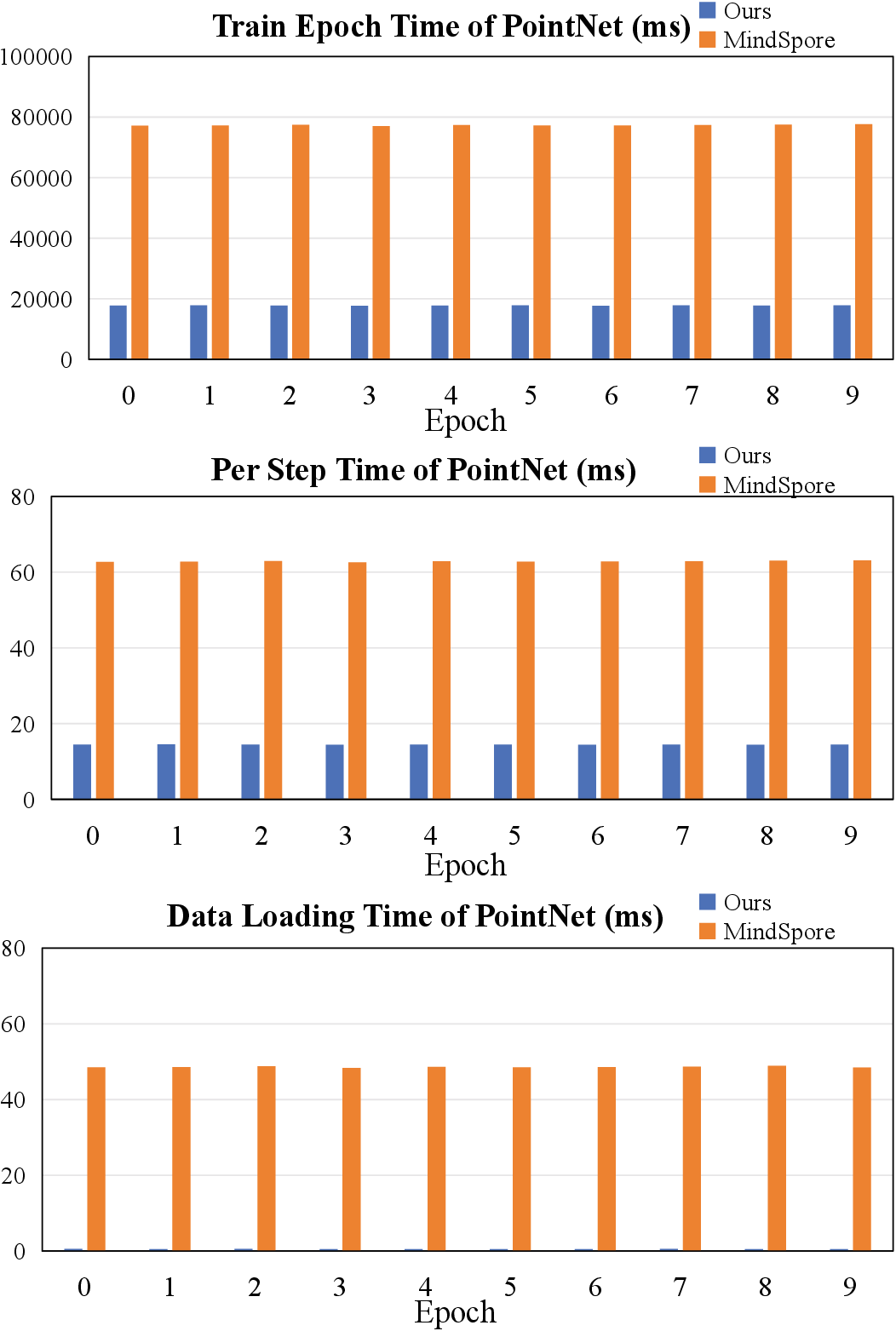}
    \caption{MindSpore vs. Ours in terms of memory usage with Dataset Kitti.}
    \label{fig:enter-label}
\end{figure}

 \begin{figure}[!h]
    \centering
    \includegraphics[width=1\linewidth]{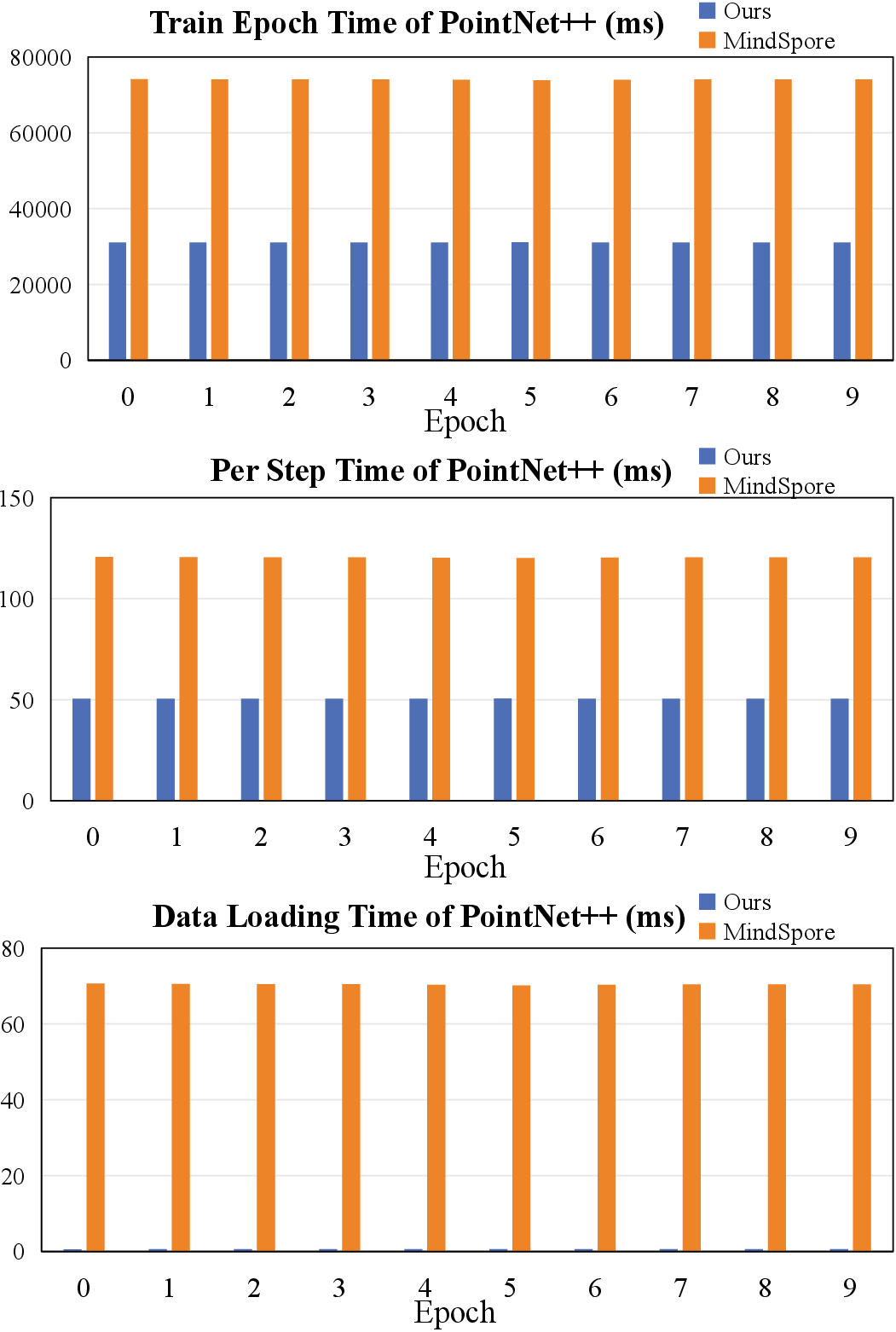}
    \caption{MindSpore vs. Ours in terms of memory usage with Dataset Kitti.}
    \label{fig:enter-label}
\end{figure}

\begin{figure}[!h]
    \centering
    \includegraphics[width=1\linewidth]{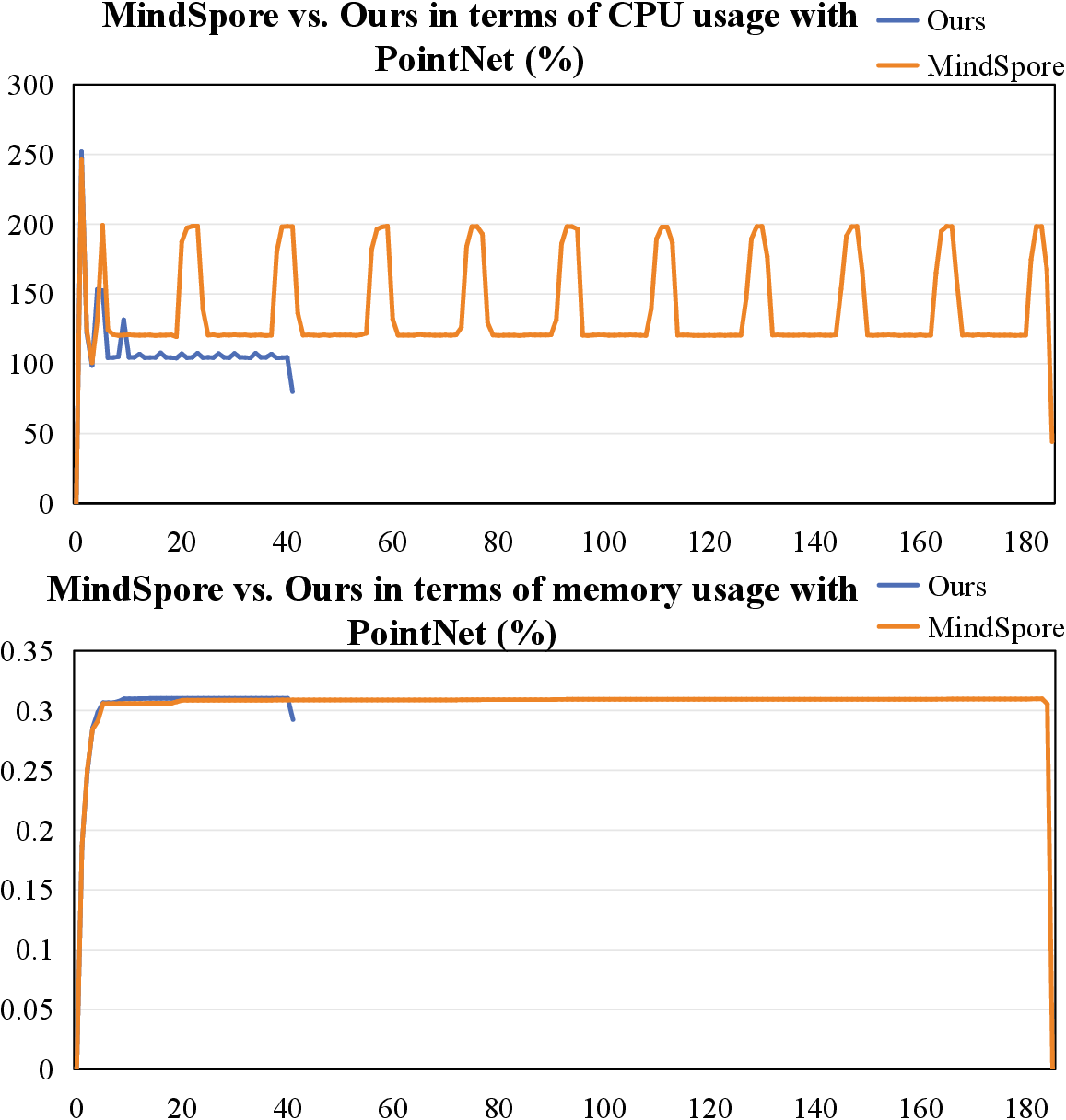}
    \caption{MindSpore vs. Ours in terms of memory usage with Dataset Kitti.}
    \label{fig:enter-label}
\end{figure}

 \begin{figure}[!h]
    \centering
    \includegraphics[width=1\linewidth]{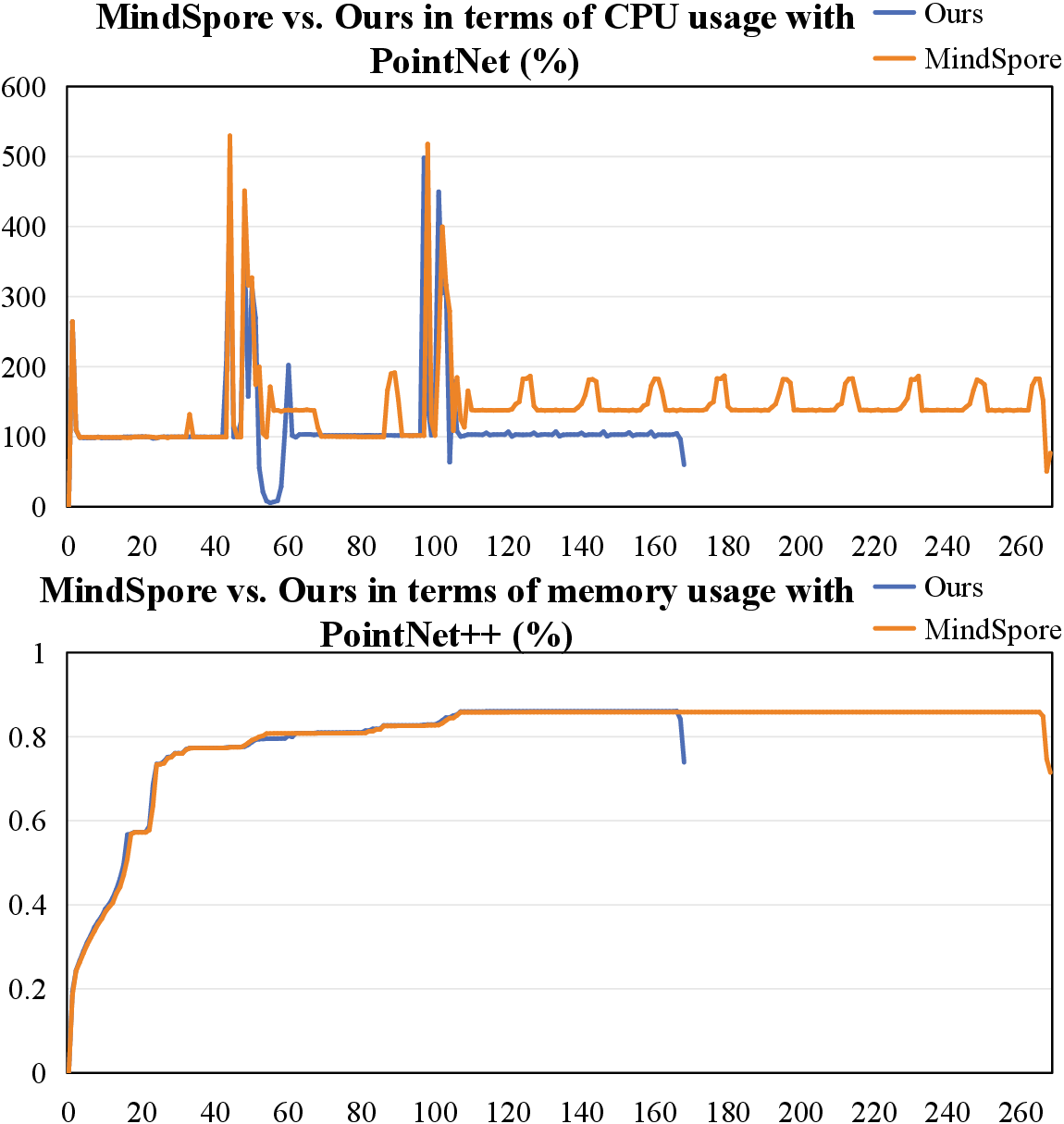}
    \caption{MindSpore vs. Ours in terms of memory usage with Dataset Kitti.}
    \label{fig:enter-label}
\end{figure}

To further validate the efficiency of our system, we conducted experiments using real-world model training tasks. Specifically, we deployed our system on the PointNet and PointNet++ models and monitored key performance metrics throughout the entire training process via dedicated monitoring threads. The results, presented in Fig. 18 and Fig. 19, demonstrate substantial performance improvements. As shown in Fig. 18, after integrating our system, the training time per epoch and per step for the PointNet model were significantly reduced, achieving an acceleration of over 4.3×, while the data loading time (including data loading and augmentation) was reduced by more than 88.5×. Similarly, for the PointNet++ model, as illustrated in Fig. 19, these two metrics improved by 2.3× and 123.5×, respectively.

These results indicate that deploying our system in model training tasks can substantially enhance the training efficiency of point cloud models by improving data throughput. By optimizing the data loading pipeline and minimizing bottlenecks in data preprocessing, our system ensures more efficient utilization of computational resources, ultimately leading to a significant reduction in overall training time.

Regarding system performance, the experimental data presented in Fig. 20 and Fig. 21 further highlight the efficiency gains achieved by our system. The results clearly demonstrate that our system significantly reduces CPU utilization, effectively alleviating computational overhead during data loading and preprocessing stages. Additionally, as the point cloud data processed in our experiments is relatively small, the memory overhead during the data processing phase remains largely unchanged. However, as dataset sizes continue to grow, the optimizations introduced by our system in data handling are expected to yield even greater benefits in terms of resource efficiency, making it highly scalable for large-scale point cloud applications.

Beyond experimental validation, our system has been successfully adopted in real-world applications. The code has been contributed to the MindSpore open-source community and is widely utilized by the MindSpore development team. Furthermore, our system has been integrated into Huawei’s point cloud processing projects, including autonomous driving technology under the Harmony Intelligent Mobility Alliance. The Harmony team employs our system to efficiently load LiDAR point cloud data, significantly enhancing performance during both training and inference stages. Additionally, the USTC\_IAT\_United team from the University of Science and Technology of China (USTC) leveraged our system for point cloud processing during their participation in The Autonomous Grand Challenge at the CVPR 2024 Workshop. Their system, powered by our data processing framework, demonstrated outstanding performance and ultimately secured first place in the competition.

These real-world deployments underscore the practical value of our system in both industrial and academic settings. Its ability to optimize point cloud data handling has proven beneficial in accelerating model training, improving computational efficiency, and supporting large-scale applications, making it a valuable tool for a wide range of point cloud processing tasks.


\nocite{*}
\bibliographystyle{IEEEtran}
\bibliography{reference}




\end{document}